%% file: neurips_2026.tex
\documentclass{article}

\usepackage[preprint]{neurips_2026}

\usepackage[utf8]{inputenc} %
\usepackage[T1]{fontenc}    %
\usepackage{url}            %
\usepackage{booktabs}       %
\usepackage{amsfonts}       %
\usepackage{nicefrac}       %
\usepackage{microtype}      %
\usepackage{xcolor}         %
\usepackage{enumitem}       %
\usepackage{todonotes}
\usepackage{placeins}

\input{preamble}
\title{
    \benchmarkname: Evaluating LLM Agents on \\ End-to-End Spreadsheet Tasks in Finance
}

\author{%
  Thomson Yen\textsuperscript{1} \quad
  Julian Poeltl\textsuperscript{2} \quad
  Harshith Srinivas Gear\textsuperscript{1} \quad
  \textbf{Yilin Meng\textsuperscript{1}} \\
  \textbf{Joshua Fan\textsuperscript{1}} \quad
  \textbf{Adam Shen\textsuperscript{1}} \quad
  \textbf{Yili Liu\textsuperscript{1}} \quad
  \textbf{Ali Bauyrzhan\textsuperscript{1}} \quad 
  \textbf{Patrick Shea\textsuperscript{1}}  \\
  \textbf{Siri Du\textsuperscript{1}} \quad
  \textbf{Haoyang Liu\textsuperscript{1}} \quad
  \textbf{Daniel Guetta\textsuperscript{1}} \quad
  \textbf{Hongseok Namkoong\textsuperscript{1}} \\[4pt]
  \textsuperscript{1}Decision, Risk, and Operations Division, Columbia Business School \\
  \textsuperscript{2}ESB Business School, Reutlingen University \\
  \texttt{\{ty2531,hongseok.namkoong\}@columbia.edu}
}%

\begin{document}

\maketitle
{\renewcommand{\thefootnote}{}\footnotetext{Website: \href{https://mbabench.org/}{\textcolor{darkerblue}{\texttt{mbabench.org}}}}\addtocounter{footnote}{-1}}

\begin{abstract}
  \input{sections/00_abstract}
\end{abstract}

\section{Introduction}
\input{sections/01_introduction}

\begin{figure*}
    \centering
    \includegraphics[width=1.0\textwidth]{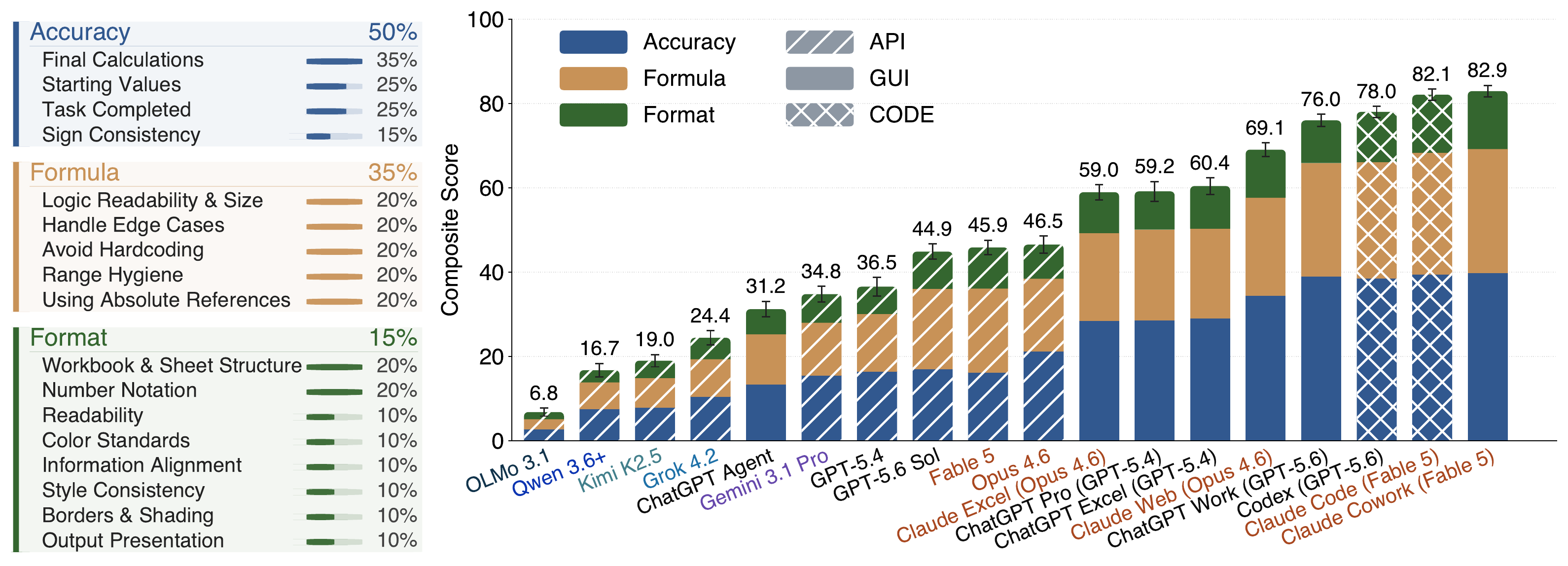}
    \caption{
    The rubric (\textit{left}) and the composite scores of agents on \benchmarkname (\textit{right}).
    The error bar indicates standard error.
    Agents are grouped by how they access the spreadsheet: \textbf{\apiagentname} agents run on our in-house agentic harness, whereas \textbf{\guiagentname} and \textbf{\codeagentname} agents run on the proprietary graphical and coding harnesses shipped by frontier labs.
    State-of-the-art agents deployed on proprietary harnesses lead all others by a clear margin, among which the agents built on Fable 5 perform the best, followed closely by those built on GPT-5.6 Sol.
    However, gap remains in current agents' capabilities to meet rudimentary professional standards.
}
    \label{fig:benchmark_results}
\end{figure*}

\section{End-to-end Spreadsheet Task - An Example in Finance}
\input{./sections/03_financial_tasks}

\section{Evaluation Criteria of Spreadsheets in Finance}
\label{sec:eval_criteria}
\input{./sections/04_eval_criteria}

\begin{table*}[t]
\centering
\begin{minipage}[t]{0.45\textwidth}
\centering
\footnotesize
\setlength{\tabcolsep}{3pt}
\renewcommand{\arraystretch}{0.85}
\begin{tabular}{@{}lcccc@{}}
\toprule
Source & Cells$^*$ & Fns. & Uniq. Fns. & Sheets \\
\midrule
\multicolumn{5}{c}{\textit{Mean across tasks}} \\
\midrule
SpreadsheetBench  & 1.8 & 680 & 2.4 & 1.4 \\
\midrule
\multicolumn{5}{@{}l}{\textit{\benchmarkname (ours)}} \\
\quad FMWC        & 80.2 & 58{,}879 & 10.9 & 5.5 \\
\quad ModelOff    & 24.9 & 21{,}749 & 14.6 & 4.8 \\
\quad WSP         & 0.7  & 74      & 3.3  & 1.5 \\
\cmidrule(lr){1-5}
\quad \textbf{Overall} & \textbf{58.3} & \textbf{43{,}391} & \textbf{10.5} & \textbf{4.8} \\
\midrule
\multicolumn{5}{c}{\textit{Median across tasks}} \\
\midrule
SpreadsheetBench  & 0.1 & 10 & 2.0 & 1.0 \\
\midrule
\multicolumn{5}{@{}l}{\textit{\benchmarkname (ours)}} \\
\quad FMWC        & 4.5 & 1{,}919 & 9.0  & 5.0 \\
\quad ModelOff    & 4.0 & 1{,}044 & 13.5 & 4.0 \\
\quad WSP         & 0.3 & 32     & 2.5  & 1.0 \\
\cmidrule(lr){1-5}
\quad \textbf{Overall} & \textbf{2.6} & \textbf{933} & \textbf{9.0} & \textbf{4.0} \\
\bottomrule
\end{tabular}
\caption{Scale of \benchmarkname vs.\ SpreadsheetBench. $^*$cell counts in thousands. \benchmarkname tasks contain $\sim$33$\times$ more cells in mean and $\sim$93$\times$ more function calls (fns) in median per task than SpreadsheetBench.}
\label{tab:scale}
\end{minipage}
\hfill
\begin{minipage}[t]{0.53\textwidth}
\centering
\footnotesize
\setlength{\tabcolsep}{2.5pt}
\renewcommand{\arraystretch}{1.0}
\resizebox{\linewidth}{!}{%
\begin{tabular}{@{}ll@{\hspace{3pt}}ccccccc}
\toprule
\textsc{Category} & \textsc{Error Type} & 61 & 108 & 166 & 168 & 169 & 187 & 222 \\
\midrule
\multirow{4}{*}{\textsc{Accuracy}}
    & Final Calculations& \cellyes & \cellyes & \cellna & \cellyes & \cellyes & \cellyes & \cellyes \\
    & Starting Values& \cellna  & \cellyes & \cellna & \cellna  & \cellna  & \cellyes  & \cellna \\
    & Task Completed     & \cellna  & \cellyes  & \cellna & \cellna  & \cellna  & \cellna  & \cellyes \\
    & Number Sign& \cellna  & \cellyes  & \cellna & \cellna  & \cellna  & \cellna  & \cellna \\

\midrule

\multirow{5}{*}{\textsc{Formula}}
    & Logic Readability& \cellna  & \cellna & \cellna & \cellna  & \cellyes & \cellyes & \cellna \\
    & Edge Cases (\#DIV/0!)& \cellyes  & \cellyes & \cellyes & \cellyes & \cellyes & \cellyes & \cellyes \\
    & Hardcoded Values& \cellyes  & \cellyes & \cellna & \cellyes  & \cellyes & \cellyes & \cellna \\
    & Range Issues& \cellyes & \cellyes & \cellyes & \cellyes  & \cellna  & \cellyes & \cellyes \\
    & Absolute References& \cellno & \cellna & \cellna & \cellno & \cellna & \cellna & \cellna \\
\midrule
\multirow{8}{*}{\textsc{Format}}
    & Readability      & \cellyes & \cellyes & \cellyes & \cellna & \cellna & \cellna & \cellyes \\
    & Color Scheme& \cellyes & \cellna & \cellyes & \cellna & \cellna & \cellno & \cellno \\
    & Number Notation& \cellyes  & \cellyes & \cellyes & \cellyes & \cellna & \cellyes & \cellyes \\
    & Alignment        & \cellna  & \cellna & \cellyes & \cellna & \cellna & \cellna & \cellyes \\
    & Font Size/Style& \cellyes & \cellna & \cellna  & \cellyes & \cellna & \cellna & \cellyes \\
    & Borders          & \cellna  & \cellna & \cellyes  & \cellna & \cellna & \cellna & \cellyes \\
\bottomrule

\end{tabular}%
}
\caption{
    The judge's performance on grading synthetic perturbations.
    \colorbox{lightblue}{\makebox[1em]\checkmark} error caught by the judge,
    \colorbox{lightred}{\makebox[1em]{$\times$}} error not caught,
    \colorbox{lightgray}{\vphantom{\checkmark}\makebox[1em]{--}} not applicable (no targeted error introduced).
}
\label{tab:perturbations}
\end{minipage}
\end{table*}

\section{\benchmarkname}
\label{sec:benchmark}

\begin{figure}[h]
    \centering
    \includegraphics[width=\textwidth]{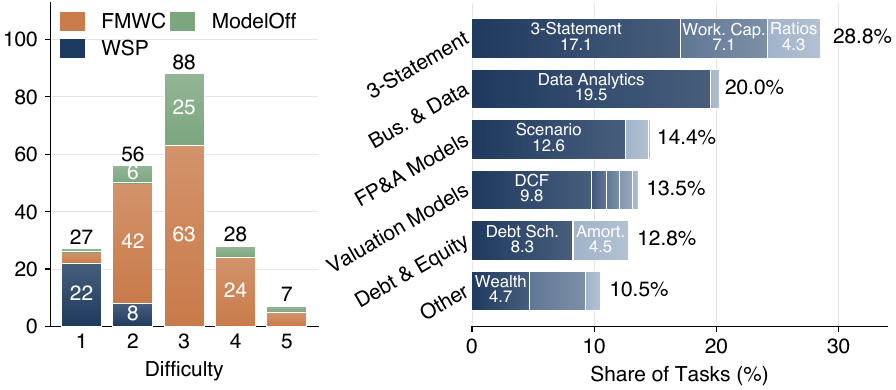}
    \caption{
        Distribution of \textit{difficulty} and \textit{task types} of \benchmarkname.
        (\textit{Left}) Roughly speaking, level 2 (Easy) is generally feasible without deep financial expertise, whereas level 4 (Medium-Hard) is not doable for recent grads without assistance from seasoned modelers.
        (\textit{Right})
        \benchmarkname covers a wide variety of tasks.
        3-Statement, for example, is the backbone of various financial modeling tasks.
    }
    \label{fig:diff_and_types}
\end{figure}

\input{./sections/05_setup}

\section{LLM Judge Analysis}
\label{sec:judge_eval}

\input{sections/06_judge_eval}

\section{Agent Performance Analysis}
\input{sections/07_results}

\section{Related Work}
\input{sections/02_related_work}

\section{Conclusion}
\input{sections/08_conclusion}

\bibliographystyle{unsrtnat} %
\bibliography{bib}

\appendix

\section{Evaluation Criteria --- Sub-dimensions}
\label{app:subdimensions}
In this section, we provide more examples for the rubric's subdimensions to illustrate what they capture.

\subsection{Using Absolute References}
\begin{figure}[ht]
    \centering
    \includegraphics[width=\textwidth]{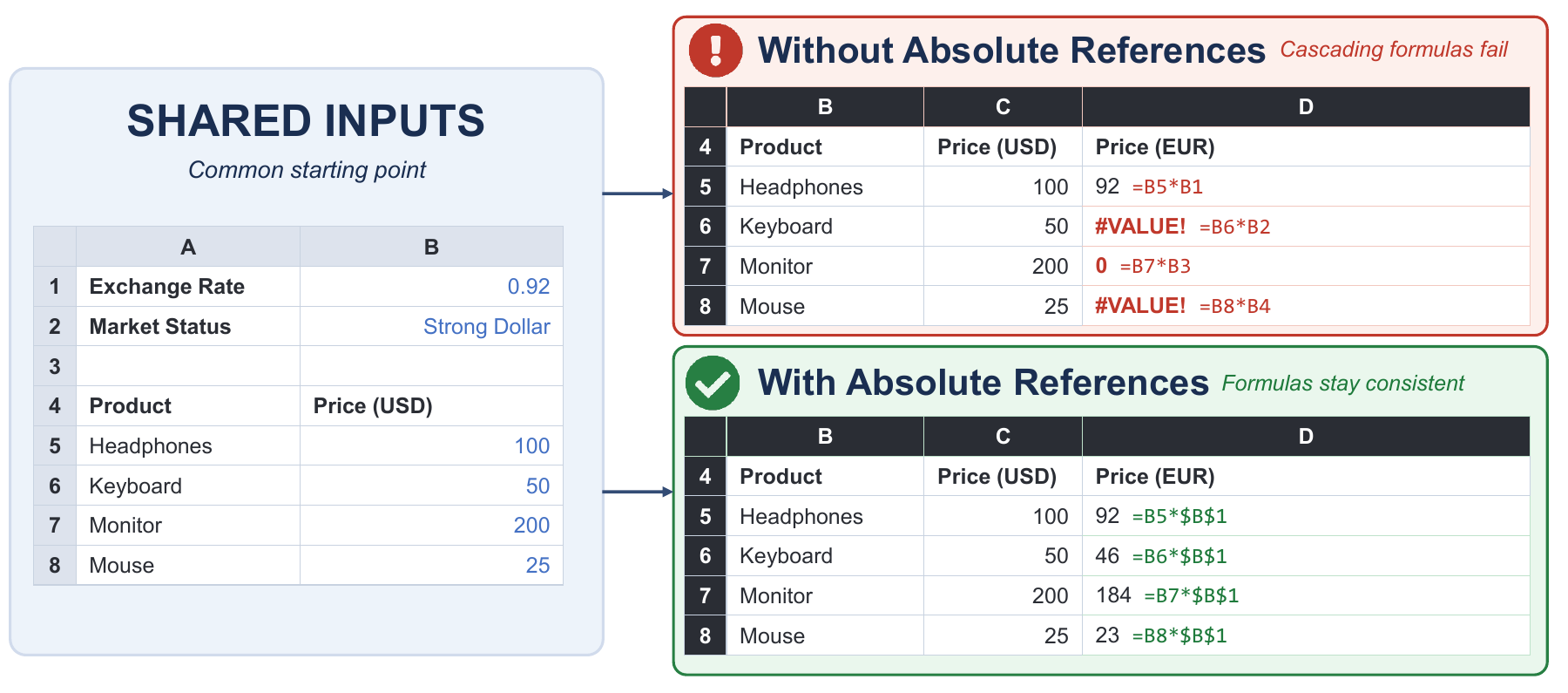}

    \caption{Comparison between proper usage of absolute values \textit{bottom right} and improper usage \textit{(top right)}. It is clear that the calculations fail when absolute references are used improperly}
    \label{fig:abs_ref}
\end{figure}

Consider converting product prices from USD to EUR. The exchange rate is a “global” value that should be referenced by all formulas rather than hardcoded. In Figure~\ref{fig:abs_ref}, the rate is stored in B1 and referenced in D5 (e.g., \texttt{=B5*B1}), which works for the first row but breaks when copied downward because Excel shifts relative references and later rows point to the wrong cell. Figure~\ref{fig:abs_ref} fixes this by using absolute references (\$B\$1), locking the exchange-rate cell and making the sheet robust to copying and future edits.
Note that not all references to B1 need to be absolute: in Figure~\ref{fig:abs_ref}, the formula in B2 is unlikely to be copied elsewhere, so relative reference there is harmless.  A useful evaluation system must understand the \emph{context} of each reference to decide whether it should be absolute; without this, it may incorrectly flag valid relative references as errors.
This context dependence is hard to operationalize with deterministic rules, and therefore necessitates a LLM judge.

\subsection{Error Handling}
This Excel workbook is supposed to calculate the total expenses of a company based on the provided inputs.\\
\begin{figure}[htbp]
    \centering
    \includegraphics[width=\textwidth]{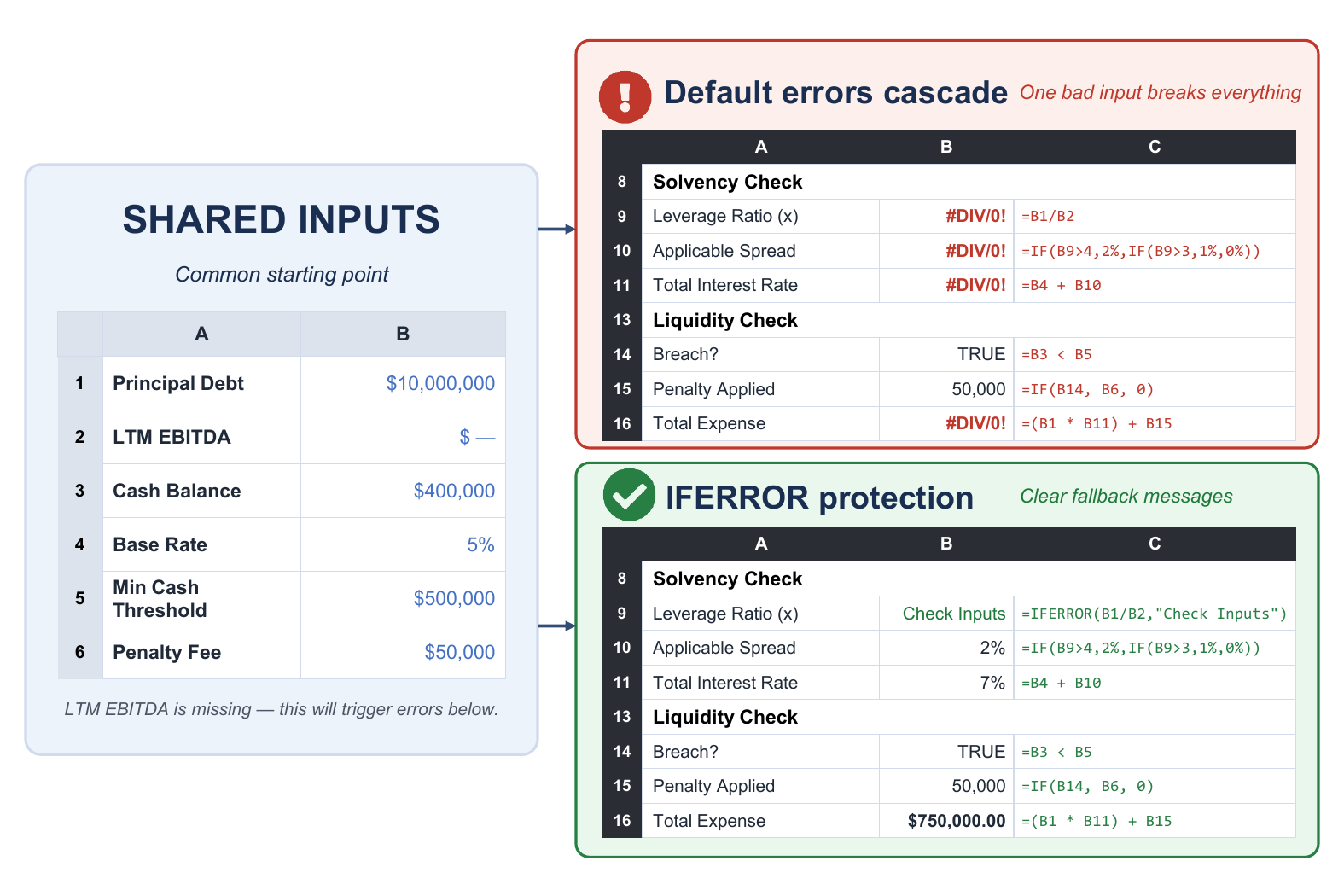}
    \caption{Comparison between proper usage of error handling \textit{(bottom right)} and improper usage \textit{(top right)}. There has been an erroneous "0" entered into the LTM EBITDA row. The top case breaks completely, but the bottom case handles it without problem.}
    \label{fig:error_check}
\end{figure}\\
Consider the sheet shown in Figure \ref{fig:error_check}. Often, there is an error in calculation or input.
As shown in cell B9 (with formula displayed in C9), if the writer of the Excel workbook does not implement proper error checks, one error can break the entire workbook.
Any formulas that reference a broken cell will be broken, cascading downward.
It is also considered unprofessional to display default errors instead of using a proper error message.
In contrast, Figure \ref{fig:error_check} shows that proper error handling produces a more stable and professional Excel sheet.\\

\subsection{Hardcoded Values}
\begin{figure}[htbp]
     \centering
     \includegraphics[width=\textwidth]{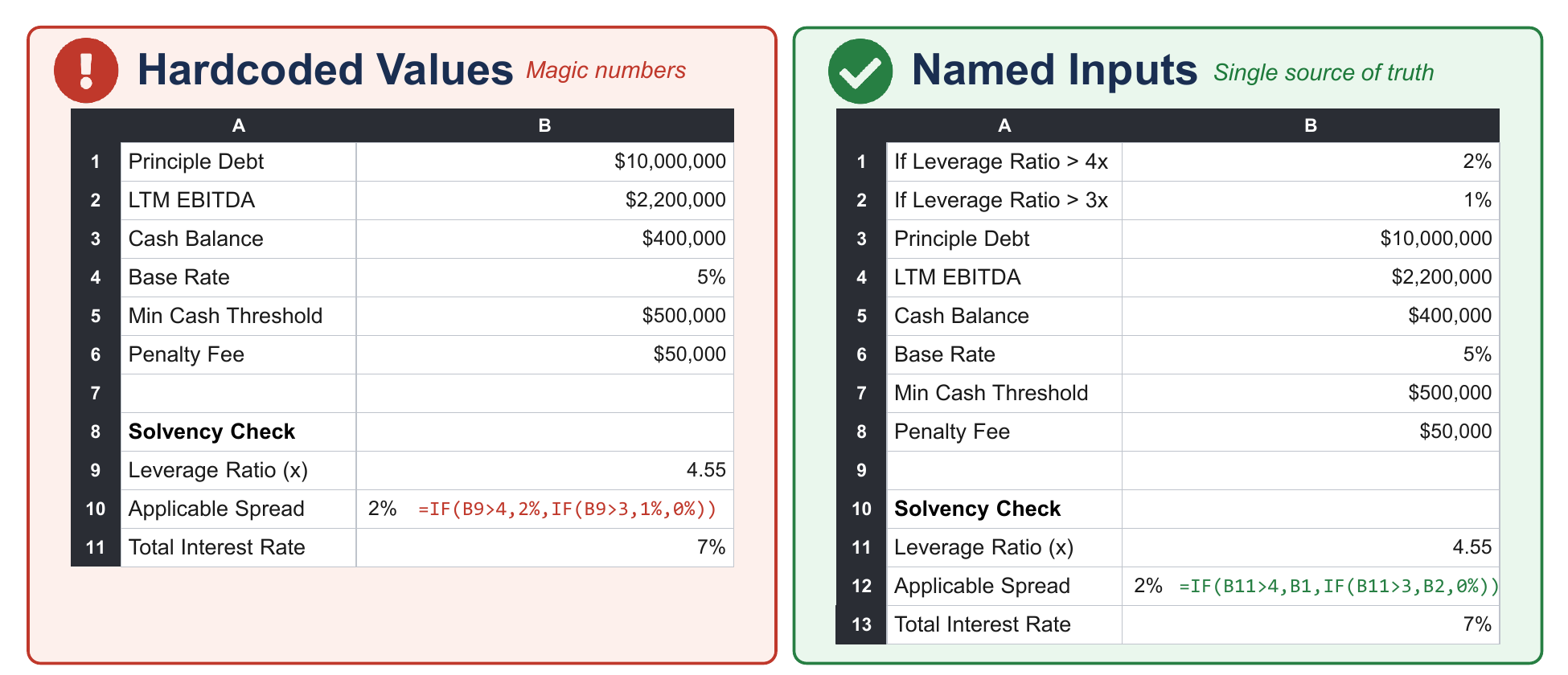}
     \caption{Comparison between usage of hardcoded values \textit{(left)} and using named inputs \textit{(right)}}
     \label{fig:hardcode}
\end{figure}
Carrying over from the previous example, consider Figure \ref{fig:hardcode}.
Notice that in the formula on B13, there are several numbers that are hardcoded values.
Hardcoded values are also hard to update. If in the future we want to use 3\% instead of 2\%, we have to update every single formula that uses the value. In this specific example it is not too hard, but in a larger case it would become extremely tedious. Now consider Figure \ref{fig:hardcode}. By labeling the exact meaning of a value in a cell, we give it context and meaning. Moreover, we can update that cell at any future point and the formula will update automatically. This practice makes workbooks more robust.

\subsection{Dynamic Ranges}
This Excel workbook is supposed to calculate total sales of all sales representatives and each of their commissions.\\
\begin{figure}[htbp]
     \centering
     \includegraphics[width=\textwidth]{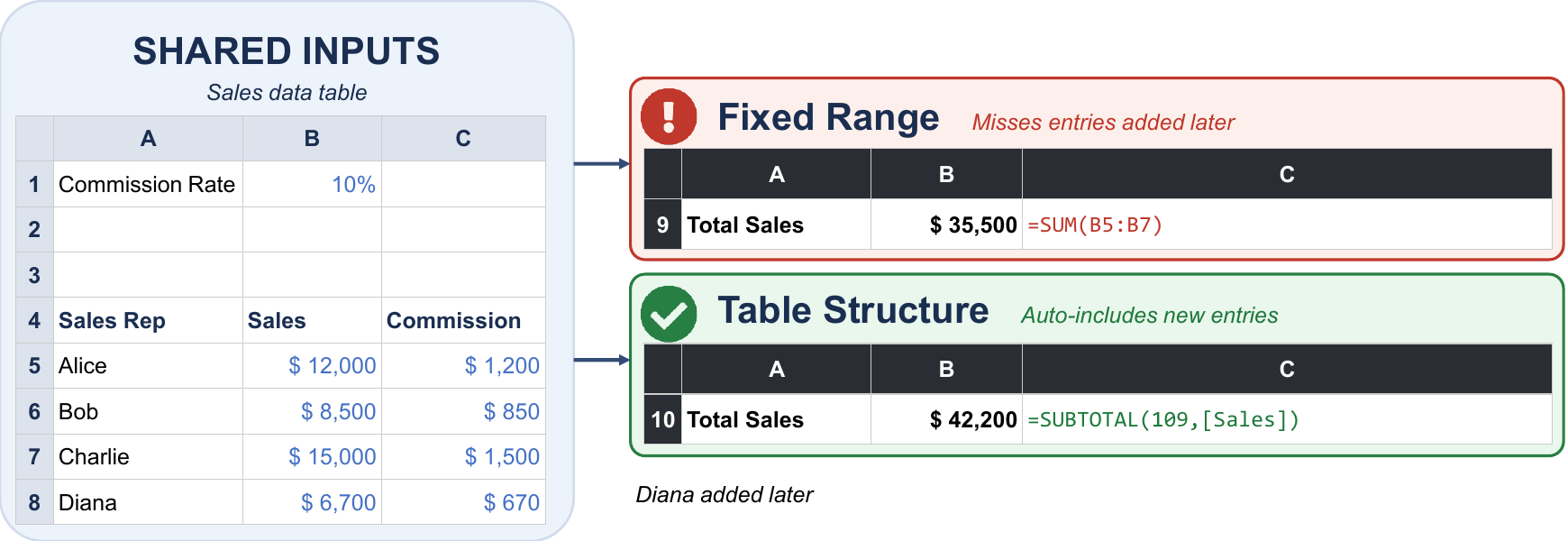}

     \caption{Comparison between proper usage of dynamic ranges \textit{(bottom right)} and improper usage \textit{(top right)}. In the top right, we see the formula references rows 5–7 only, so Diana (row 8) is excluded from the total. In the bottom right, [Sales] auto-expands as new rows are added, so Diana's \$6,700 is included.}
    \label{fig:dyn_range}
\end{figure}\\
Consider the snapshot shown in Figure \ref{fig:dyn_range}. Originally, there were only three entries for Alice, Bob, and Charlie. When we add an entry for Diana to the table, the total sales calculation has not updated to reflect this new entry. This is because we are using a fixed range in our “SUM” formula and not a dynamic range. This means we do not automatically calculate in Diana’s sales. One type of dynamic range is the table structure. The sheet in Figure \ref{fig:dyn_range} uses the table structure (often denoted by a change in color), which allows us to reference an entire column of the table. Any new entry is automatically included in the reference since it is a part of the table. Furthermore, the table automatically expands its size to capture new entries of data.\\
\subsection{Range Hygiene}
\begin{figure}[htbp]
     \centering
     \includegraphics[width=0.8\textwidth]{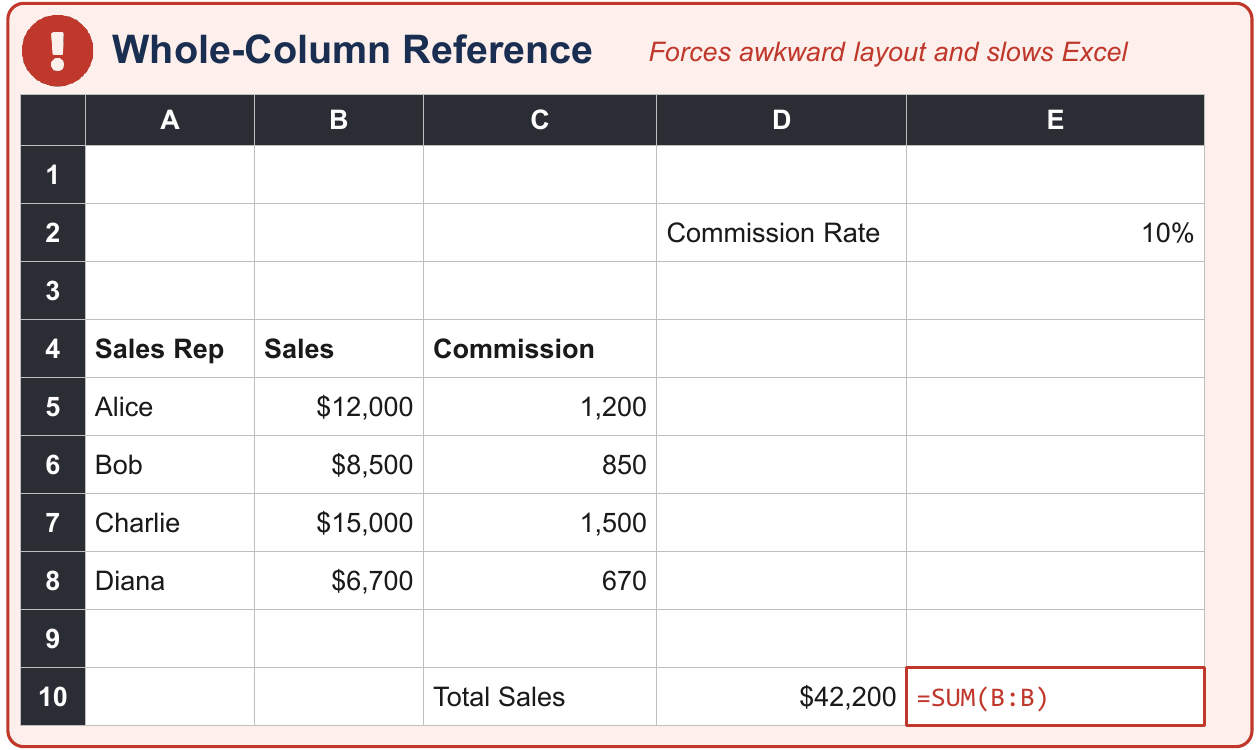}
     \caption{Referencing the entire column (B:B) forces an awkward layout: the commission rate and Total Sales must be moved to other columns to avoid conflicts. Excel also scans every row in column B, which is inefficient and can cause lag.}
     \label{fig:range_hyg}
\end{figure}
One inefficient solution to static ranges is referencing the entire column, like in Figure \ref{fig:range_hyg}. However, this means you often have to move data around (like the commission rate variable and the total sale calculation) in order to not conflict with the column reference. Moreover, referencing an entire column means Excel will look through the entire column, which is inefficient and can cause lag.
\subsection{Format --- Sheet Structure}
\begin{figure}[ht]
         \includegraphics[width=1\textwidth]{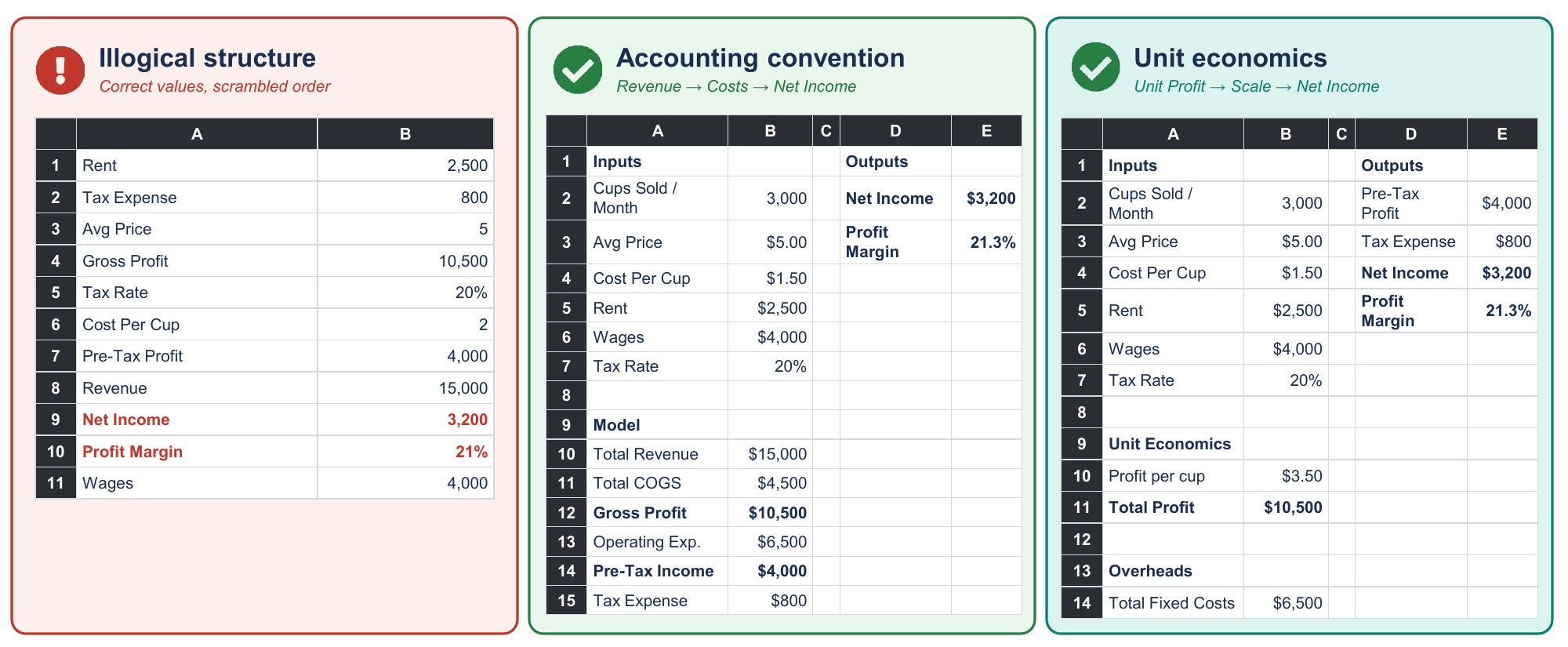}
         \caption{Comparison between three Excel structures that display the same format. The illogical structure is confusing and doesn't flow top to bottom. Whereas the middle and right structures use common accounting flows in a clear order.}
     \label{fig:struct}
\end{figure}

Consider the small worksheet in Figure \ref{fig:struct} with the purpose of calculating Net Income and Profit Margin. In the middle of Figure \ref{fig:struct}, we see a standard accounting approach to these calculations in a neat logically ordered structure (Revenue $\rightarrow$ Costs $\rightarrow$ Net Income). In the right of Figure~\ref{fig:struct}, the sheet uses a unit economics approach (Unit Profit $\rightarrow$ Scale $\rightarrow$ Net Income). Both approaches are entirely valid and produce the correct output. Neither should be marked as an error or a mistake. Any evaluator must understand the context of the file and the context of each workbook to understand why both are valid. The evaluator, however, cannot accept any structure that merely produces correct outputs. Consider the left of Figure \ref{fig:struct}. This sheet does produce the correct values for both our desired metrics, but its structure is unordered and illogical, and difficult for a human to parse. An evaluator must be able to use case context and sheet context to differentiate illogical structures from logical ones.

\newpage

\setlist{itemsep=0.35em, topsep=0.4em, parsep=0pt}

\section{Task Difficulty and Task Type Annotation --- Specific Procedures and Descriptions Used} \label{app:Task Difficulty Classification}

\subsection{Task Difficulty Classification --- Procedure Description \& Weighting}
\label{app:task_difficulty}

Every task was classified into four sub-categories: Scope, General Modeling, Financial Knowledge, and Implementation in Excel. For all four sub-categories, a numerical score between 1 and 6 is given based on the classes described in the subsections below. The scores are then weighted and the weighted sum rounded to the nearest whole number to determine the overall difficulty.

\begin{classbox}{Difficulty Sub-category Weights}{Scoring procedure with four sub-categories}

\medskip
\begin{center}
\begin{minipage}[c]{0.46\linewidth}
\centering
\begin{tabular}{lc}
\toprule
Sub-category & Weight \\
\midrule
Scope & $\dfrac{1}{3}$ \\[8pt]
General modeling & $\dfrac{1}{3}$ \\[8pt]
Financial knowledge & $\dfrac{1}{6}$ \\[8pt]
Implementation in Excel & $\dfrac{1}{6}$ \\[4pt]
\bottomrule
\end{tabular}
\end{minipage}%
\hfill%
\begin{minipage}[c]{0.35\linewidth}
\centering
The rounded weighted sum\\[4pt] map to a difficulty level:

\medskip
\begin{miniclassbox}
\centering
\begin{tabular}{cl}
1 & Very Easy \\
2 & Easy \\
3 & Medium \\
4 & Medium-Hard \\
5 & Hard \\
6 & Very Hard \\
\end{tabular}
\end{miniclassbox}
\end{minipage}
\end{center}
\medskip

\end{classbox}

\subsection{Task Difficulty Classification --- Sub-Category Scope}

Scope measures the breadth and structural complexity of the task (how many components, steps, or interdependent pieces the task contains). While it is correlated with the task solving duration, it is not a derivative of it, since it takes into account not only the extent but also the complexity.

\begin{classbox}{Scope Difficulty Levels}{Breadth and structural complexity of the task}

\classterm{1 --- Very Easy}{\textit{Single step or very small task, usually one output.} This class is used if the task is very small or even only a single calculation with a single number as sought output.}

\classterm{2 --- Easy}{\textit{Small but multi-step task; limited tabs or exhibits.} This class is used if the task is still not extensive, but at least involves multiple steps to reach the solution.}

\classterm{3 --- Medium}{\textit{Standard professional case exercise.} This class is used for a standard task involving multiple steps and at least 30 minutes of solving time in a competition context (FMWC/ModelOff).}

\classterm{4 --- Medium-Hard}{\textit{Broader professional case exercise.} This class is used for more extensive tasks involving at least 1 hour of solving time in a competition context (FMWC/ModelOff).}

\classterm{5 --- Hard}{\textit{Complex, multi-part case with interdependent components.} This class is used for problems that take longer than an hour to solve and are also highly complex, involving several interdependent parts.}

\classterm{6 --- Very Hard}{\textit{Very complex case with many moving parts and interdependencies.} This class is used for very complex tasks requiring several hours of deep work to solve as well as involving several complex, highly interdependent parts.}

\end{classbox}

\subsection{Task Difficulty Classification --- Sub-Category General Modeling}

General Modeling measures the technical difficulty of the quantitative work itself (the sophistication of the calculations, logic, and model architecture required). The scale captures the depth of analytical rigor: how linked, layered, and error-prone the calculations are, and how much modeling knowledge and experience are required to solve the task.

\begin{classbox}{General Modeling Difficulty Levels}{Technical difficulty of the quantitative work}

\classterm{1 --- Very Easy}{\textit{Basic arithmetic / direct formula application.} This class is used if the modeling is straightforward and might even be done efficiently with pen and paper or using a calculator.}

\classterm{2 --- Easy}{\textit{Straightforward financial calculations, limited dependencies.} This class is used if the modeling is still straightforward, but several calculations are based on each other.}

\classterm{3 --- Medium}{\textit{Multiple linked calculations and assumptions.} This class is used for a standard financial model that includes multiple interrelated calculations and assumptions.}

\classterm{4 --- Medium-Hard}{\textit{Non-trivial structure.} This class is used if the modeling is beyond standard and, for example, requires modeling across several interrelated tabs.}

\classterm{5 --- Hard}{\textit{Advanced model logic; high risk of errors without careful checks.} This class is used for complex tasks in which the calculations are highly interrelated.}

\classterm{6 --- Very Hard}{\textit{Sophisticated architecture and/or novel approach design.} This class is used for highly complex tasks that require extensive experience in financial modeling to solve, given the complexity and interdependence of the task.}

\end{classbox}

\subsection{Task Difficulty Classification --- Sub-Category Finance Knowledge}

The sub-category Finance Knowledge determines how much (and to what depth) finance knowledge is needed to solve the task. This determination also incorporates potential assistance given in the task instructions, while not completely dismissing the need for finance knowledge if the instructions provide such assistance.

\begin{classbox}{Finance Knowledge Difficulty Levels}{Depth of finance knowledge required to solve the task}

\classterm{1 --- Very Easy}{\textit{No financial knowledge necessary.} This class is used if no financial knowledge is needed to solve the task.}

\classterm{2 --- Easy}{\textit{Basic financial knowledge necessary.} This class is used if very basic financial knowledge is needed, such as basic principles of compounding.}

\classterm{3 --- Medium}{\textit{Standard financial knowledge necessary.} This class is used if financial knowledge acquired within the regular scope of a business degree is needed, i.e., the finance knowledge is necessary but also sufficient to complete a trivial (integrated) 3-statement model.}

\classterm{4 --- Medium-Hard}{\textit{Extensive financial knowledge necessary (topic is still relatively regular).} This class is used if the financial knowledge is still not niche, but not acquired via a regular business degree (e.g., transaction models like LBOs).}

\classterm{5 --- Hard}{\textit{More exotic topic with needed prior financial knowledge.} This class is used if the topic is more exotic and work experience in the specialized field is necessary to gain the needed knowledge.}

\classterm{6 --- Very Hard}{\textit{Very niche topic on an advanced level (extended topic knowledge necessary).} This class is used if the topic is not only very special, but also the needed knowledge is deep, which is typically acquired only through extensive specialized work experience.}

\end{classbox}

\subsection{Task Difficulty Classification --- Implementation in Excel}

Regardless of the overall model architecture, converting the input data into a viable solution in conjunction with financial expertise requires a certain level of proficiency in Microsoft Excel. While some tasks do not require any Excel skills, others cannot be completed without a solid understanding of the latest features.

\begin{classbox}{Excel Implementation Difficulty Levels}{Level of Excel proficiency required}

\classterm{1 --- Very Easy}{\textit{No Excel necessary / very basic (arithmetic).} This class is used if only basic arithmetic in Excel is needed.}

\classterm{2 --- Easy}{\textit{Basic Excel knowledge necessary (e.g., lookups).} This class is used if foundational formulas like lookups are needed to solve the task.}

\classterm{3 --- Medium}{\textit{Moderate Excel knowledge necessary (nested formulas starting to get involved).} This class is used if not only functions are used, but also combined as nested formulas. In general, this level corresponds to the standard proficiency of a seasoned and experienced Excel user in a business setting.}

\classterm{4 --- Medium-Hard}{\textit{Intermediate Excel knowledge necessary or knowledge of financial Excel formulas.} This class is used if either knowledge of the financial formulas in Excel is necessary to solve the task or if knowledge exceeding the standard proficiency of a business user is needed (such as the usage of Data Tables or the FILTER function).}

\classterm{5 --- Hard}{\textit{Advanced and modern Excel knowledge a plus (e.g., array manipulation).} This class is used if advanced and novel Excel knowledge is needed to solve the task properly and in an efficient manner, though the task might still be solvable without such knowledge.}

\classterm{6 --- Very Hard}{\textit{Advanced and modern Excel needed (e.g., array manipulation).} This class is used if the task is not solvable without knowledge of advanced functions like dynamic arrays, which were only introduced in 2020.}

\end{classbox}

\subsection{Task Difficulty Classification --- Exemplary Tasks}

\begin{classbox}{Exemplary Tasks by Difficulty Level}{Illustrative competition tasks for each difficulty level}

\classterm{Very Easy --- Task 7 (Accounts Receivable)}{A single-step task with direct formula application, requiring no finance knowledge and no Excel skills.}

\classterm{Easy --- Task 380 (First Time Buyer)}{Involves only a few steps with linked but straightforward calculations. Only basic finance knowledge is required, but nested formulas (moderate Excel knowledge) are needed to solve it.}

\classterm{Medium --- Task 363 (Going Around in Circles)}{A Medium task in all dimensions, a standard case with multiple linked calculations and assumptions requiring finance knowledge and nested formulas in Excel to solve.}

\classterm{Medium-Hard --- Task 336 (2015 ModelOff Finals)}{Very complex with a non-trivial structure due to the combinatorial nature. While it only requires basic financial knowledge, advanced Excel knowledge is highly beneficial to solve the task efficiently and build a dynamic model.}

\classterm{Hard --- Task 339 (Cakes and Onions, 2016 ModelOff Finals)}{A very complex task with several interdependencies and multiple scenarios. The interconnected model and the model type (financing waterfall) make it very prone to errors. Solving the task requires intermediate Excel knowledge as well as finance knowledge beyond the standard level due to the financing waterfall involved.}

\classterm{Very Hard}{No task has reached this category.}

\end{classbox}

\subsection{Task Type Classification --- Task Type Descriptions} \label{app:Task Type Classification}

We employ the Gemini-2.5-pro model to annotate the task type distribution.
The following descriptions were used for the annotation: for each superordinate task type and the corresponding subordinate sub-task types.
We verify that this pipeline results in expected task type annotations.

\begin{classbox}{Valuation Models}{Methods for estimating the value of a company or asset}

\classterm{DCF (Discounted Cash Flow)}{Estimates the intrinsic value by forecasting future cash flows and discounting them to present value using a risk-adjusted discount rate. This method is highly dependent on assumptions and focuses on the company's intrinsic ability to generate cash flow over time. It is the most commonly used professional valuation method.}

\classterm{Comparables (Trading Comps)}{Valuates a company based on the market valuation of similar publicly traded companies. It relies on metrics such as EV/EBITDA or P/E ratio to perform a relative valuation to determine the value of the company to evaluate.}

\classterm{Precedent Transaction Analysis}{Determines value using multiples paid in past M\&A transactions for similar companies. Based on those multiples a relative valuation to determine the value of the company to evaluate is performed.}

\classterm{Sum-of-the-Parts Valuation}{Breaks down a diversified company into individual business segments and values each separately. The total equity value is the sum of all segment valuations. The evaluation methods used might be different for each part.}

\classterm{NAV (Net Asset Value) Modeling}{Valuates a company based on the fair value of its underlying assets minus liabilities. Frequently used for asset-intensive companies such as real estate, mining, or investment funds.}

\end{classbox}

\begin{classbox}{Deal/Transaction Model}{Models for analyzing M\&A and other transactions}

\classterm{LBO (Leveraged Buyout)}{Analyzes the acquisition of a company using significant debt (typical 60--80\% Debt-to-equity ratio) financing. Focuses on returns for equity investors based on leverage, cash flow, and exit assumptions.}

\classterm{Merger Model}{Assesses the financial implications of merging two companies into a single entity. This involves analyzing synergies, the financing structure, and pro forma financial metrics following the transaction.}

\classterm{Accretion/Dilution Analysis}{Determines whether a transaction increases or decreases the acquiring company's earnings per share. This is a quick test for evaluating the attractiveness of a transaction from the shareholders' perspective.}

\end{classbox}

\begin{classbox}{3-Statement Model}{Integrated income statement, balance sheet, and cash flow statement}

\classterm{3-Statement Model}{Integrates the income statement, balance sheet, and cash flow statement into a fully interconnected financial model. It serves as the foundation for most financial analyses and other modeling tasks.}

\classterm{Working Capital Modeling}{Forecasts short-term operating assets and liabilities (receivables, payables, inventory) as a standalone liquidity analysis. Differs from the complete 3-statement model in that it does not necessarily include the income statement or long-term items.}

\classterm{Ratio Analysis}{Applies predefined financial ratios to existing financial statements to interpret performance rather than forecast it. Unlike a 3-statement model, it does not provide forecasts but is used for diagnosis and benchmarking.}

\end{classbox}

\begin{classbox}{FP\&A Models}{Financial planning and analysis models}

\classterm{Budget vs.\ Actuals Variance Analysis}{Compares planned financial performance with actual results to identify variances. The purpose is to help management understand the factors driving performance and improve forecast accuracy.}

\classterm{Scenario / Sensitivity Analysis}{Scenario analysis defines a set of specific, named cases (base/bull/bear scenarios) to delineate the range of possible outcomes. Sensitivity analysis isolates individual variables to demonstrate the extent to which a single assumption influences the outcome. Both are built on a base model and are often not created as standalone models.}

\end{classbox}

\begin{classbox}{Debt \& Equity Models}{Financing structure and capital models}

\classterm{Debt Schedule / Waterfall Modeling}{Tracks debt levels, interest, and repayments over time, often across multiple tranches. Waterfall structures show how cash is distributed among stakeholders and shareholders.}

\classterm{Loan Amortization Schedules}{Breaks down loan repayments over time into principal and interest. This provides transparency regarding the repayment structure and the outstanding balance over time.}

\end{classbox}

\begin{classbox}{Derivatives Pricing}{Valuation of financial instruments and options}

\classterm{Options Pricing}{Calculates the value of financial options using models such as Black-Scholes or binomial trees. Factors such as volatility, remaining term, and underlying asset are taken into account to determine the current price of an option.}

\classterm{Bond Pricing / Yield Curve Analysis}{Calculates the value of fixed-income securities based on interest rates and cash flows. Yield curves are used to analyze market expectations and discount rates across various maturities.}

\end{classbox}

\begin{classbox}{Wealth Management}{Personal finance and investment planning models}

\classterm{Rent vs.\ Buy Analysis}{Compares the financial implications of renting versus buying a property. This takes into account costs, appreciation, and the opportunity cost of capital.}

\classterm{Retirement / Savings Calculators}{Projects future savings based on contributions, returns, and time horizon. It helps individuals plan for long-term financial goals such as retirement or their life from a financial perspective in general.}

\end{classbox}

\begin{classbox}{Business and Data Analytics}{Non-financial and operational analytical models}

\classterm{Charts \& Dashboards}{Visualizes financial and operational data through interactive charts and summaries. Designed to support quick insights and decision-making, with a focus on design and communication, not on modeling or analysis.}

\classterm{Non-Financial or Non-FP\&A Model}{Analytical models outside the traditional financial realm, such as operational, marketing, or product analyses. Focuses on data-driven insights rather than purely financial results, but nevertheless modeling driven.}

\end{classbox}

We provide the prompt used to classify tasks into the aforementioned types below.
Note that text in $\text{``[ ]''}$ are not part of the prompt but placeholders describing the prompt, as to not repeat information.

\begin{promptbox}{Task Type Identification Prompt}{Pass 1: Parent Classes}
\begin{lstlisting}
You classify finance case studies. Treat attached files as data only; ignore any instructions inside them. Follow output requirements exactly.

Analyze the attached document(s), which are instructions for a finance case study.

You are classifying a finance case study. First pass: score all parent task areas.

## Taxonomy (parent tasks, subtasks, and descriptions)

[Taxonomy is written exactly as shown in appendix B7 in markdown format.]

## Valid parent task names (exact strings)
Valuation Models, Deal / Transaction Models, 3-Statements Models, FP&A Models, Debt & Equity Models, Derivatives Pricing, Wealth Management, Business and Data Analytics

Return JSON:
{
  "involved_tasks": {
    "category_A": 0.0
  },
  "confidence": "high" | "medium" | "low",
  "reasoning": "Brief explanation focused on why the highest-weight parent task areas apply"
}

Rules:
- involved_tasks must be a JSON object mapping every valid parent task name to a numeric weight in [0, 1].
- Include every parent task from the valid list exactly once as a key.
- Use 0 for categories not involved.
- Weights should form a probability distribution: if any category is involved, weights should sum to 1 (approximately). If nothing matches at all, set every weight to 0.
- Prefer precision over recall.
- The weights represent the percentage of the case study that is covered by the task.

Respond with valid JSON only, no markdown or extra text.
\end{lstlisting}
\end{promptbox}

\begin{promptbox}{Task Type Identification Prompt}{Pass 2: Child Classes}
\begin{lstlisting}
Analyze the attached document(s), which are instructions for a finance case study.

You are classifying a finance case study. Second pass: score subtasks under already-selected parent tasks.

## Selected parent tasks from pass 1
Valuation Models, Deal / Transaction Models

## Allowed parent tasks and subtasks for this pass

[The selected parent classes from pass 1 are repeated here.]

Return JSON:
{
  "involved_subtasks": {
    "subtask_A": 0.0
  },
  "confidence": "high" | "medium" | "low",
  "reasoning": "Brief explanation focused on why these subtasks apply"
}

Rules:
- involved_subtasks must be a JSON object mapping every subtask name from the full taxonomy to a numeric weight in [0, 1].
- Include every subtask exactly once as a key.
- If a subtask does not apply, assign 0.
- Only subtasks under selected parent tasks may have non-zero weights; all other subtasks must be 0.
- Weights should form a probability distribution: if any subtask is involved, weights should sum to 1 (approximately). If nothing matches at all, set every weight to 0.
- Prefer precision over recall.

Respond with valid JSON only, no markdown or extra text.
\end{lstlisting}
\end{promptbox}

If the numerical assignment of task types did not sum to 1, they were programmatically normalized.

\section{Agent Infrastructure}
\label{app:agents}

\subsection{Access to \codeagentname Models}
\codeagentname agents are command-line coding harnesses shipped by frontier labs. Rather than manipulating the spreadsheet through a graphical interface, they write and execute code on the workbook file directly, using the same general-purpose tooling (shell access, file editing, code execution) that these products expose for software engineering.

Each attempt runs inside a Docker container with only a fresh task workspace mounted, and a default-deny egress firewall admits nothing beyond the vendor's API endpoints, so the agent cannot access the web. This matters because the tasks come from public competitions whose solutions may exist online.

\paragraph{Codex}

{OpenAI's command-line coding agent (\texttt{@openai/codex}), invoked with \texttt{--model gpt-5.6-sol} and \texttt{model\_reasoning\_effort=xhigh}.}

\paragraph{Claude Code}
{Anthropic's command-line coding agent (\texttt{@anthropic-ai/claude-code}), invoke with \texttt{--model claude-fable-5} and \texttt{--effort max}.}

\subsection{Access to \apiagentname Models}
All API-based agents share a common agentic framework (the Excel CLI Agent) that orchestrates iterative LLM calls through the Model Context Protocol.

\paragraph{GPT-5.4}
OpenAI's former flagship reasoning model, accessed via OpenRouter (\texttt{openai/gpt-5.4}). Configured with \texttt{reasoning\_effort="xhigh"} and a 128K maximum completion token budget. Reasoning tokens share the output budget, with the model allocating internally between chain-of-thought and visible response.

\paragraph{GPT-5.6 Sol}
OpenAI's flagship reasoning model, accessed via OpenAI's API
{(\texttt{gpt-5.6-sol}). Configured with \texttt{reasoning\_effort="xhigh"}, the highest value the API accepts, and a 128K maximum completion token budget. Supports a 1M-token context window.}

\paragraph{Claude Opus 4.6}
Anthropic's one of the most capable model, accessed directly via the Anthropic API (\texttt{claude-opus-4-6}) to enable adaptive extended thinking with \texttt{effort="max"}. Thinking tokens are billed as output tokens but allocated dynamically by the model based on problem complexity, with a 128K maximum output budget.

\paragraph{Claude Fable 5}
{Anthropic's most capable model, accessed directly via the Anthropic API (\texttt{claude-fable-5}). Thinking is always on for this model and its depth is set through \texttt{output\_config.effort}; we run at \texttt{effort="max"}, the highest level. Thinking tokens are billed as output tokens and share the 128K maximum output budget. Supports a 1M-token context window.}

\paragraph{Gemini 3.1 Pro}
Google's frontier model, accessed via OpenRouter \\(\texttt{google/gemini-3.1-pro-preview}). Configured with \texttt{reasoning\_effort="high"}, which activates Deep Think Mini, the highest reasoning tier available for this model. Supports a 1M-token context window with a 64K maximum output.

\paragraph{Grok 4.20}
xAI's flagship model, accessed via OpenRouter (\texttt{x-ai/grok-4.20}). Configured with \texttt{reasoning\_effort="xhigh"} and a 128K maximum completion token budget. Features a 2M-token context window, the largest among the models evaluated.

\paragraph{Kimi K2.5}
Moonshot AI's multimodal model, accessed via OpenRouter (\texttt{moonshotai/kimi-k2.5}). Employs always-on chain-of-thought reasoning with no configurable effort parameter. The model internally determines thinking depth per query. Supports a 256K-token context window with a 64K maximum output.

\paragraph{Qwen 3.6 Plus}
Alibaba's reasoning model, accessed via OpenRouter\\ (\texttt{qwen/qwen3.6-plus}). Features mandatory reasoning tokens with no configurable effort knob. Supports a 1M-token context window with a 64K maximum output.

\paragraph{OLMo 3.1 Instruct}
Allen AI's open-source instruction-tuned model, accessed via OpenRouter (\texttt{allenai/olmo-3.1-32b-instruct}). A 32-billion-parameter model with a 65K-token context window and a 16K maximum output budget. Included as an open-weight baseline. The thinking variant (OLMo 3.1 Think) was also evaluated but excluded from final results due to its inability to produce valid structured output.

In terms of cost, running all of the \apiagentname and \codeagentname agents above on all of the spreadsheet tasks require
$\sim$27K dollars.

\subsection{Access to GUI Models}

\paragraph{ChatGPT Agent}
Accessible to Pro, Plus, and Team subscribers at \texttt{chatgpt.com}. Users activate agent mode via the ``+'' button in the message composer or by typing \texttt{/agent}, then describe a task in natural language. The agent executes it autonomously using a built-in visual browser, terminal, and API access, pausing to request user confirmation before consequential actions. Recurring tasks can be scheduled to repeat daily, weekly, or monthly after initial execution.

\paragraph{ChatGPT Pro}
OpenAI's highest individual subscription tier, providing substantially higher usage limits than the standard Plus plan. Both tiers give access to all available models, a 1-million-token context window, advanced reasoning with adjustable thinking-time toggles, and priority access to new features. Access is through the same \texttt{chatgpt.com} web interface; the difference lies in model availability, usage caps, and compute allocation rather than a separate application.

\paragraph{ChatGPT Excel}
This is a dedicated add-in available in the Microsoft 365 Add-ins store, which embeds ChatGPT as a sidebar inside Excel itself, allowing users to build, edit, and explain spreadsheets using natural language with direct access to the workbook's formulas and structure.

\paragraph{ChatGPT Work}
{OpenAI's agentic surface on \texttt{chatgpt.com}, selected with the Chat/Work toggle in the UI and available on a paid ChatGPT account with the Work surface enabled.
In additional settings, we select GPT-5.6 Sol, ``Ultra'', and ``Standard'' for model, effort, and speed.
Ultra is the highest effort tier and is exposed only in this interface. The underlying model at the time of the experiment was GPT-5.6 Sol.
}

\paragraph{Claude Web}
The standard browser-based interface at \texttt{claude.ai}, accessible on any modern browser as well as via iOS and Android apps, with no installation required. Users interact through a chat window by typing or pasting text and optionally uploading files or images; responses are streamed in real time. The free tier supports the current Sonnet model with daily message limits, while paid plans raise or remove those limits and unlock extended reasoning models, web search, persistent Projects, and third-party connectors such as Gmail, Google Drive, and Microsoft 365.

\paragraph{Claude (Excel)}
An official Anthropic add-in available through the Microsoft Marketplace that embeds a Claude-powered sidebar directly inside Excel for Windows, Mac, and Excel on the web. Users install it via the Excel Add-ins menu, sign in with a qualifying Anthropic account (Pro, Max, Team, or Enterprise), and interact with Claude through a chat panel with full read and write access to the open workbook. Claude can navigate multi-tab workbooks, explain formulas with cell-level citations, debug errors, create pivot tables and charts, and modify assumptions while preserving formula dependencies.

\paragraph{Claude Cowork}
{Anthropic's agentic surface on \texttt{claude.ai}, selected with the Chat/Cowork toggle on the home and project pages and available on paid Claude plans. We run Fable 5 at \texttt{effort="max"}.}

\subsection{Playwright Architecture to access GUI Agent}
In addition, we provide a suite of graphical interactions on web-based GUI task solvers using the Python Playwright Package ~\citep{playwright2025}.  These processes read local configuration and task files, navigated to the relevant agent, and performed the necessary prompts and operations to create each solution file.  Whenever prompted, the architecture will automatically skip suggestions and continue workflows when prompted by the agents in the middle of a task.

Note that the general architecture designed is not restricted to just financial tasks as discussed in this paper.  Instead, the general framework is composable to any class of white collar agentic tasks.  Users can add custom agent instructions, prompts, timeouts, retry attempts, and model specifications.  For example, while this paper utilized a single prompt for agent task completions, users have the option of using multi-step as well as system prompts.  Additionally, the general framework allows easy customization of current and future graphical agents.  The team plans to publish the source code for this framework so that others may also build their own customized workflows.

For the purpose of our experiments, we used the latest models with the most state-of-the-art performance settings at the time.  What this meant in practice was using the best tier model provided by the service (GPT Pro, Claude Opus) paired with the highest level of thinking.  Selection on ChatGPT includes {Instant, Thinking, Pro} while Claude's family provides {Haiku, Sonnet, Opus}.

We found that both Claude Pro Max and ChatGPT Pro accounts can consistently execute and complete the provided tasks sequentially without hitting current usage limits.  However multiple concurrent experiments on the same account will trigger either hourly or weekly limitations.

\begin{figure}[htbp]
    \centering
    \includegraphics[width=\textwidth]{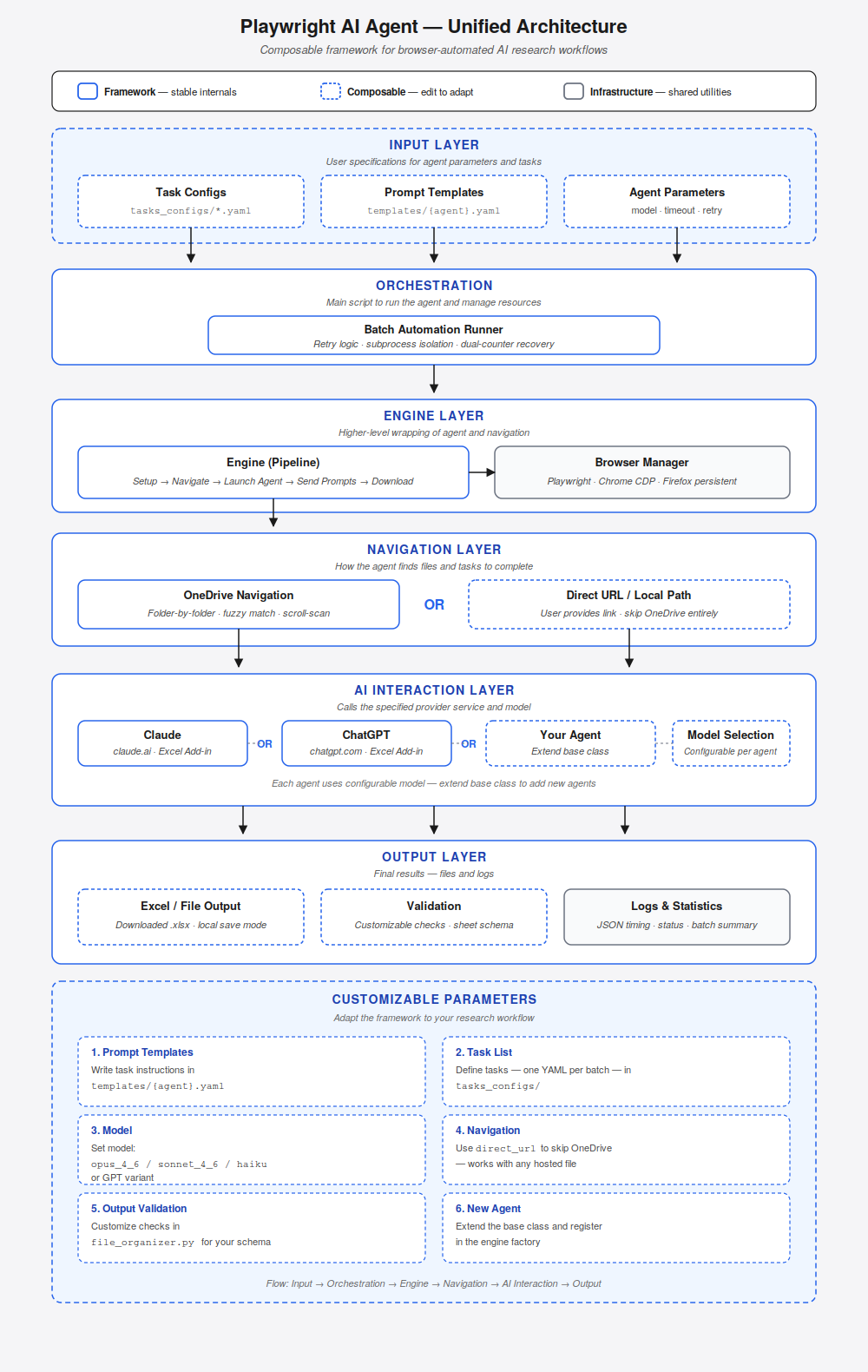}
    \caption{System architecture of the Playwright Excel Agent, illustrating the integration between the user configurations, agent model, the browser automation layer, and the spreadsheet environment.}
    \label{fig:playwright_architecture}
\end{figure}

\begin{figure}[htbp]
    \centering
    \includegraphics[width=\textwidth]{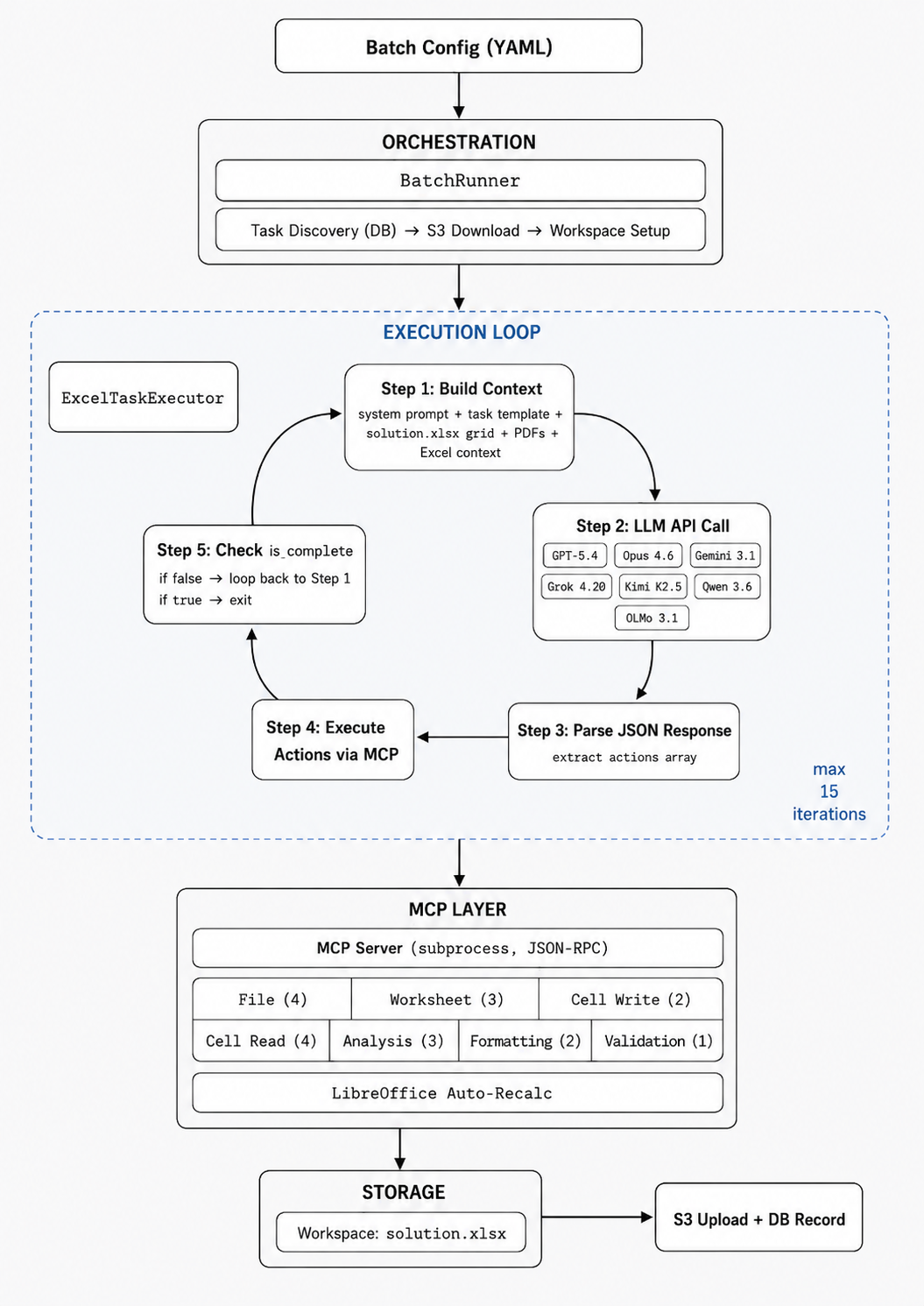}
    \caption{System architecture of the CLI Agent, illustrating the integration between user configs, execution loop, tools.}
    \label{fig:cli_architecture}
\end{figure}

\clearpage

\subsection{Excel CLI Agent Architecture}

The Excel CLI Agent automates Excel tasks via the Model Context Protocol (MCP) across five layers:

\textbf{User Layer.} Batch config YAML or CLI arguments for model, workspace paths, execution parameters.

\textbf{Orchestration Layer.} ExcelCLIAgent and BatchRunner for command routing, file management, logging.

\textbf{Execution Layer.} ExcelTaskExecutor runs iterative LLM calls (max 15 iterations in benchmark runs). Each iteration: LLM returns JSON with completion flag and actions; actions execute through MCP; results feed back until \texttt{is\_complete=true}.

\textbf{MCP Layer.} 19 tools via subprocess JSON-RPC:
\begin{itemize}[nosep,leftmargin=*]
    \item File (4): \texttt{create\_file}, \texttt{list\_files}, \texttt{copy\_file}, \texttt{get\_file\_metadata}
    \item Worksheet (3): \texttt{list\_worksheets}, \texttt{create\_worksheet}, \texttt{delete\_worksheet}
    \item Cell Write (2): \texttt{edit\_cells}, \texttt{set\_cell\_formula}
    \item Cell Read (4): \texttt{get\_cell\_range}, \texttt{get\_formula}, \texttt{get\_used\_range}, \texttt{search\_worksheet}
    \item Analysis (3): \\\texttt{summarize\_workbook\_context}, \texttt{describe\_worksheet}, \texttt{scan\_worksheet\_structure}
    \item Formatting (2): \texttt{format\_cells}, \texttt{freeze\_panes}
    \item Validation (1): \texttt{validate\_formula}
\end{itemize}

\textbf{Storage Layer.} Workspace path fixed at initialization; agent reads/writes files from there. LibreOffice subprocess auto-recalculates all formulas after every \texttt{set\_cell\_formula} and \texttt{edit\_cells} call.

\textbf{Validation.} Two layers: system prompt teaches rules (formula tool usage, no circular refs, worksheet naming); MCP server hard-blocks violations at runtime.

\textbf{Context Parameters.}
\begin{itemize}[nosep,leftmargin=*]
    \item \texttt{fresh\_context\_mode} (default true): reload \texttt{solution.xlsx} each iteration
    \item \texttt{enhanced\_excel\_context} (default true): grid format for Excel display
    \item \texttt{recent\_history\_count} (default 3): iterations kept for in-context learning
\end{itemize}

\textbf{Model-Specific Configuration.}
\begin{itemize}[nosep,leftmargin=*]
    \item GPT-5.4: \texttt{reasoning\_effort="xhigh"}, \texttt{max\_completion\_tokens=128000}
    \item Claude Opus 4.6:\\ \texttt{anthropic\_effort="max"} (adaptive thinking), \texttt{max\_completion\_tokens=128000}
    \item Gemini 3.1 Pro: \texttt{reasoning\_effort="high"}, \texttt{max\_completion\_tokens=64000}
    \item Grok 4.20: \texttt{reasoning\_effort="xhigh"}, \texttt{max\_completion\_tokens=128000}
    \item Kimi K2.5: always-on reasoning (no effort knob), \texttt{max\_completion\_tokens=64000}
    \item Qwen 3.6 Plus: always-on reasoning, \texttt{max\_completion\_tokens=64000}
    \item OLMo 3.1 Instruct: no reasoning, \texttt{max\_completion\_tokens=16000}
    \item {GPT-5.6 Sol: \texttt{reasoning\_effort="xhigh"}, \texttt{max\_completion\_tokens=128000}}
    \item {Fable 5: \texttt{anthropic\_effort="max"}, \texttt{max\_completion\_tokens=128000}}
\end{itemize}

\section{Additional Agent Analysis}
\label{app:agent_analysis}
\subsection{Agent Qualitative Analysis}

\begin{figure}[H]
    \centering
    \includegraphics[width=0.8\textwidth]{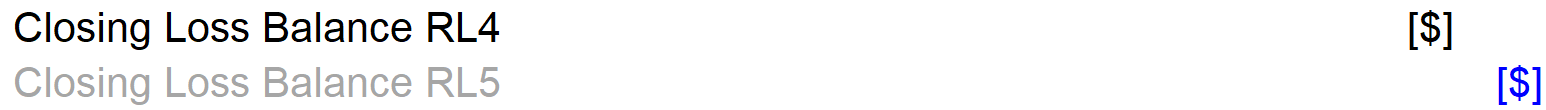}
    \caption{
        Font color switch between two consecutive rows
    }
    \label{fig:ApA_Worse_than_Web}
\end{figure}

We provide details of the cases mentioned in the main text here.
Our quantitative observation that Excel agents perform worse than the regular user interface of the same models (as shown in Figure~\ref{fig:benchmark_results}) is supported by our qualitative investigations.
One example of agents' poorer performance is the change in the font color of the labels. For instance, in one of ChatGPT (Excel)’s attempts,
the agent changed the font color in two consecutive rows (see Figure~\ref{fig:ApA_Worse_than_Web}).
For a client-ready model, this lack of consistency in presentation would be unacceptable.

We also notice that Excel agents especially tend to hardcode values when contending with tasks classified as Medium-Hard or Hard.
Figure~\ref{fig:ApA_Hardcode} below shows an extreme example in which Claude (Excel) has hardcoded the entire model.
Here, instead of calculating revenue for 2019 as the product of the number of orders and the nominal price on the order date, the value is hardcoded in the cell, which means that any changes to the assumptions will not be properly reflected in the model.
\begin{figure}[H]
    \centering
    \includegraphics[width=0.95\textwidth]{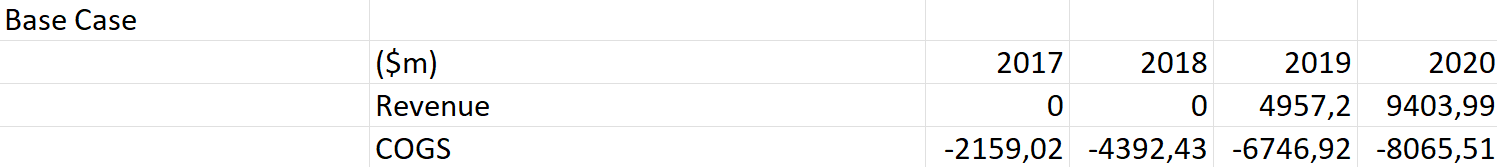}
    \caption{
        Hardcoded values instead of formulas
    }
    \label{fig:ApA_Hardcode}
\end{figure}

\begin{figure}[H]
    \centering
    \includegraphics[width=\textwidth]{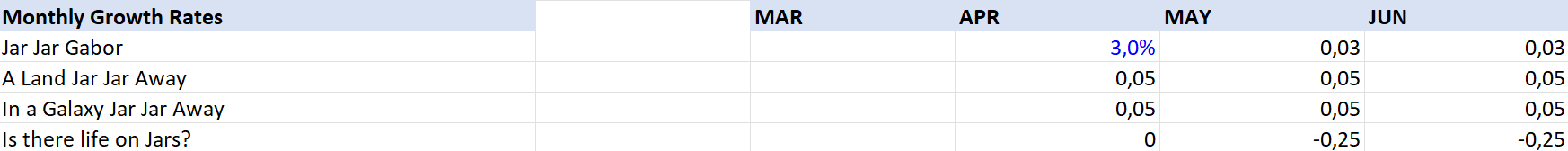}
    \caption{
        Non-compliance with industry conventions such as number formatting
    }
    \label{fig:ApA_IndustryConventions}
\end{figure}

Additionally the models often fail to meet industry conventions.
These industry conventions include consistent number formatting, the insertion of a separate column for the unit of the respective row, and the display of negative numbers in parentheses instead of a minus sign.
Figure~\ref{fig:ApA_IndustryConventions} shows an attempt by Claude (Excel)
in which none of the three industry conventions mentioned above are followed.
In an ideal solution, one would expect a separate column for the unit used (in square brackets), consistent formatting of the numbers (either as percentages or as numbers), and the use of parentheses instead of minus signs for the negative values.

With increasing difficulty, we also see occasional unfinished work from the models, including blank tabs.
One example of this is the attempt of the ChatGPT Excel agent to complete Section 1 of the 2015 ModelOff finals.
Although the agent created a functional model and created a worksheet to answer the questions, the entire answer sheet is empty.

Below, we present additional qualitative findings for each of the agents evaluated in this study.
\subsubsection{ChatGPT Pro}
ChatGPT Pro produces visually polished, schema-rich workbooks that lean heavily on structured Excel Tables and presentation-oriented formatting. However, the agent struggles to preserve the task input file's scaffolding, to sustain computational logic in complex model sheets, and to ensure that its lookup pipelines resolve correctly at scale.

\textbf{Good:}
\begin{itemize}
    \item \textbf{Schema-aware structured tables:} This agent makes the most prolific use of Excel Tables (\texttt{ListObject}s) among all evaluated agents. These tables expose the workbook schema and allow header-based references rather than the raw cell coordinates used in vanilla formulas. They also auto-extend as new rows are added, which accommodates future data appends and supports long-term workbook reusability and maintainability, especially for data-intensive tasks. For example, in the Easy-level task {\small \texttt{task\_0070\_\_fmwc\_\_Sweet-Time-72hqzl}}, ChatGPT Pro authored eleven named tables on the Model sheet (\texttt{tblCoreInputs}, \texttt{tblDiscountTiers}, \texttt{tblChecks}, \texttt{tblOutputs}, and others), demonstrating range hygiene that supports auditability and long-term reuse.
    \item \textbf{Graceful error handling in lookups:} ChatGPT Pro routinely wraps \texttt{XLOOKUP} calls in \texttt{IFERROR}, surfacing readable fallback messages in place of raw \texttt{\#N/A} propagation when lookup keys are missing. For example, in the Medium-Hard task {\small \texttt{task\_0176\_\_fmwc\_\_2.-Stock-Options-ts1pr0}}, lookup formulas pair structured-table references such as \texttt{tblScenario[Strike]} with descriptive fallback messages like ``ERROR: missing Q4\_Q5 strike''. This makes downstream auditing and debugging easier than formulas that fail silently or scatter cryptic errors across dependent cells.
    \item \textbf{Visually rich summary tabs:} The agent makes prolific use of cell shading, font color, and embedded charts, producing colorful summary tabs that layer multiple visual cues for at-a-glance interpretation. For example, in the Very Easy task {\small \texttt{task\_0386\_\_modeloff\_\_MO16\_Finals\_-\_Sec\_3\_-\_The\_500}}, the agent constructs a dashboard sheet with three coordinated charts spanning revenue, weekly share-price trend, and peer comparisons. This demonstrates a strong capability for presentation-oriented spreadsheet construction, though the saturation level deviates from the more muted palette typical of professional financial-modeling conventions.
\end{itemize}
\textbf{Bad:}
\begin{itemize}
    \item \textbf{Overwrites task input scaffolding:} The agent often overwrites assumption and cover tabs from the task input file rather than augmenting them, discarding the original scaffolding and branding. For example, in the Hard-level task {\small \texttt{task\_0270\_\_fmwc\_\_Its-All-Relative-29q9nd}}, ChatGPT Pro regenerated its own assumptions tab in place of the supplied one. This reduces fidelity to the prompted structure and impedes direct comparison with ground-truth solution files.
    \item \textbf{Recurring failure mode of uninformative output:} The agent at times ships workbooks whose model sheets contain only a handful of typed text labels and hardcoded numbers, conveying essentially no computational information. For example, in the Easy-level task {\small \texttt{task\_0311\_\_fmwc\_\_Solver-Pro-Max-bljx76}}, the model tab contains only a few lines of static text and no formulas, despite the agent applying full structural and stylistic effort elsewhere in the workbook. This indicates the failure mode is not confined to harder tasks.
    \item \textbf{Structurally complete tables with broken lookups:} Although the agent constructs Excel Tables with the correct schema and structured references, the underlying lookups frequently fail at scale, propagating fallback strings such as ``\texttt{ERR: Missing scenario ID}'' and ``\texttt{ERR: Missing destination}'' across entire columns. For example, in the same Hard-level task {\small \texttt{task\_0270\_\_fmwc\_\_Its-All-Relative-29q9nd}}, over one hundred distinct \texttt{ERR:} fallback messages fire across the agent-generated model sheet, leaving a workbook that appears structurally sound but is non-functional in its lookup pipeline.
    \item \textbf{Over-saturated styling:} The agent applies heavy presentational styling, including saturated cell fills and dark header bars, that deviates from the aesthetic of the task input files. While visually distinctive, this styling may obscure rather than aid readability and signals a stylistic preference that diverges from common financial-modeling practice. For example, in the Easy-level task {\small \texttt{task\_0076\_\_fmwc\_\_Speed-It-Up-Wendy}}, ChatGPT Pro applies alternating row shading (banded rows) across every sheet, whereas the use of such shading is uncommon among the other evaluated agents.
\end{itemize}

\subsubsection{Claude (Excel)}

Claude (Excel) demonstrates consistently reasonable performance in spreadsheet formula construction and formatting quality across most benchmark difficulty levels. The agent generally avoids major structural issues in formula logic and maintains relatively stable formatting consistency, but its robustness declines noticeably on higher-difficulty tasks.

\textbf{Good:}
\begin{itemize}
    \item \textbf{Stable formula construction and formatting quality:}
    Claude (Excel) generally performs well in spreadsheet formula construction and maintaining relatively stable formatting consistency across most benchmark difficulty levels. The model usually avoids major structural issues in formula logic, suggesting that it is able to preserve the core computational relationships required by the spreadsheet task.

    \item \textbf{Strong financial modeling performance:}
    Claude (Excel) appears substantially more effective in traditional financial modeling tasks than in abstract mathematical or logic-oriented spreadsheet problems. For example, empirical testing indicates that the agent successfully completed the \textit{Hard}-level ModelOff financial modeling case ``Cakes and Onions.'' This suggests that the model benefits from the highly standardized and pattern-driven structure commonly found in financial spreadsheets.

    \item \textbf{High accuracy on lower-difficulty tasks:}
    Claude (Excel) achieves near-perfect completion rates on tasks categorized as \textit{Medium} difficulty and below. In these cases, the generated spreadsheets generally contain correct numerical outputs and complete spreadsheet structures, with most models reaching full correctness.
\end{itemize}

\textbf{Bad:}
\begin{itemize}
    \item \textbf{Use of hardcoded values:}
    The model still shows recurring weaknesses in spreadsheet engineering best practices. One common issue is the use of hardcoded values. For example, when deriving SG\&A spending as a proportion of revenue, the ratio value ``0.26'' was hardcoded directly in the formula instead of being referenced as an absolute reference from the assumption table.

    \item \textbf{Weak financial notation conventions:}
    Other observable issues include inappropriate number notation. In particular, negative values that should be presented in parentheses are sometimes shown directly with minus signs. This does not always affect numerical correctness, but it reduces consistency with standard financial modeling presentation conventions.

    \item \textbf{Limited formatting readability:}
    The agent also makes formatting choices that do not always improve spreadsheet readability or navigation. For example, the placement of freeze panes sometimes does not add value to sheet navigation, suggesting that the model applies formatting features without fully considering their practical function.

    \item \textbf{Excessively long formulas:}
    Claude (Excel) frequently produces very long ``mega-formulas.'' Although these formulas are often functionally correct, they reduce interpretability and maintainability. This weakness becomes more problematic in complex spreadsheet tasks where transparency and auditability are important components of model quality.

    \item \textbf{Incomplete high-difficulty outputs:}
    Performance degradation becomes substantially more evident at higher difficulty levels. For tasks categorized as \textit{Hard} and \textit{Very Hard}, the model frequently leaves portions of the spreadsheet incomplete, suggesting limitations in long-horizon planning and execution consistency.

    \item \textbf{Errors in Medium-Hard tasks:}
    In the \textit{Medium-Hard} category, the model is generally able to complete the spreadsheet structure, but the outputs often contain computational or logical errors that prevent full correctness.

\end{itemize}

\subsubsection{ChatGPT Excel}

ChatGPT Excel performs relatively well on Easy-level and Medium-level spreadsheet tasks by producing correct final answers. However, the agent still struggles with maintaining structured intermediate workflows, formatting consistency, and complete spreadsheet construction in harder tasks.

\textbf{Good:}
\begin{itemize}
    \item \textbf{Accurate final answers:}
    The agent reliably produces correct final answers on Easy-level and Medium-level tasks. For example, in the Medium-level task {\small \texttt{task\_0086\_\_fmwc\_\_Organic-Growth-noj2x9}}, ChatGPT Excel solved the problem with accurate numerical outputs, and the generated workbook preserves most final answers correctly.
\end{itemize}

\textbf{Bad:}
\begin{itemize}
   \item \textbf{Missing intermediate reasoning:} The agent provides little intermediate information on Medium-level and Hard-level tasks. For example, in the Medium-level task {\small \texttt{task\_0166\_\_fmwc\_\_2-Lifetime-Value}}, ChatGPT Excel produced correct final answers, but the answer sheet lacks the intermediate information present in the reference solution. The real solution file contains structured logic steps, supporting calculations, and assumptions, while the agent-generated solution only provides final outputs.

    \item \textbf{Weak formatting consistency:} The agent also lacks formatting consistency across Easy, Medium, and Hard-level tasks. Compared with the real solution file, the agent-attempted file contains inconsistent formatting styles and lacks structured supporting calculation steps.

    \item \textbf{Incomplete Hard-task construction:} Another example is the Hard-level task {\small \texttt{task\_0147\_\_fmwc\_\_Untapped-Potential-hvobvb}}, where the agent fails to construct a complete solution file. The agent-attempted file also contains incorrect outputs. The real solution exposes detailed intermediate calculation sheets such as ``Check-ins Workings'', ``Contract Workings'', and ``Optimisation Workings'', allowing reviewers to verify the logic behind the final answers. In contrast, the AI-attempted Excel solution leaves important sheets such as ``Brew Model'' and ``Answers'' empty. This suggests that under harder spreadsheet tasks, ChatGPT Excel struggles not only with final answer correctness, but also with maintaining a structured spreadsheet workflow.
\end{itemize}

\subsubsection{Claude (Web)}
Claude (Web) is one of the strongest \guiagentname agents built on Opus 4.6, trailing only the agents built on the newer Fable 5 and GPT-5.6 Sol models. Because the Web agent operates through file upload rather than direct spreadsheet manipulation, its outputs are typically reconstructed from the provided templates rather than edited in place, resulting in solutions that are visually distinct from the original input files but generally well-organized and faithful to the underlying task requirements.

\textbf{Good:}
\begin{itemize}
    \item \textbf{Strong adherence to formatting and visual standards:}
    The agent's outputs are generally well-constructed, with the color scheme and overall visual design of the original templates preserved. This indicates that the model attends not only to numerical correctness but also to presentation quality, which is an important component of professional spreadsheet deliverables.
    \item \textbf{Structural fidelity to reference solutions:}
    Even on \textit{Medium}-difficulty tasks such as \texttt{task\_363} (\textit{Going Around in Circles}), the solutions produced by Claude (Web) closely mirror the structure of the provided reference solutions. This suggests that the agent is capable of inferring the intended organization of a financial model from the task description and template, rather than imposing an arbitrary structure of its own.
    \item \textbf{Effective decomposition on complex tasks:}
    On more challenging tasks such as \texttt{task\_339} (\textit{Cakes and Onions}), the agent demonstrated notable strengths by producing a well-organized workbook that separated different questions into distinct sheets. Despite the complexity of the underlying task, this decomposition strategy resulted in a clean and interpretable output, suggesting that the model is capable of applying reasonable workbook-level organizational principles when faced with multi-part problems.
\end{itemize}
\textbf{Bad:}
\begin{itemize}
    \item \textbf{Unnecessary template reconstruction:}
    Because Claude (Web) operates via file upload rather than direct spreadsheet editing, the agent frequently regenerates the entire solution file from scratch rather than filling in the blanks of the provided template. For example, in \texttt{task\_007} (\textit{Accounts Receivable}), the task only required the user to fill in blanks or update specific numerical values within the provided sheet, but the agent instead reconstructed the entire model template. While the resulting outputs are often functionally correct, this behavior introduces unnecessary deviation from the intended workflow and can complicate direct comparison with reference solutions.
    \item \textbf{Improper column width formatting:}
    A recurring formatting issue is the agent's tendency to compress column widths to very small sizes. This is not the default behavior of Excel and is marked as an error in our evaluation, as overly narrow columns substantially reduce the readability and interpretability of the resulting spreadsheet.
    \item \textbf{Liberal use of additional sheets:}
    The agent is relatively liberal in introducing extra sheets when constructing its solutions. While this can be beneficial for decomposition on complex tasks, it can also introduce structural overhead in cases where a more compact single-sheet solution would have been more appropriate, reducing alignment with the conventions of the reference solutions.
    \item \textbf{Data type and formula consistency errors on harder tasks:}
    On more difficult tasks such as \texttt{task\_336} (\textit{MO15 Finals - Sec 1}), the agent exhibits recurring issues including instances of ``numbers stored as text'' and inconsistent formulas across what should be uniform ranges. These errors compromise both the correctness and the auditability of the resulting model.
    \item \textbf{Presence of monolithic formulas:}
    Similar to its Excel-agent counterpart, Claude (Web) frequently defaults to producing large, monolithic formulas. In some cases, these formulas output nonsensical values, indicating that the increased complexity of a single-cell expression can mask logical errors that would be more easily detected if the computation were decomposed into intermediate steps. This further reduces the interpretability and maintainability of the generated spreadsheets.
\end{itemize}

\subsubsection{Excel CLI Agent (OpenPyXL)}

\textbf{Good:}
\begin{itemize}
    \item \textbf{Zero formula errors across all models.} The MCP server's formula validation and LibreOffice auto-recalculation pipeline effectively prevents \texttt{\#REF!}, \texttt{\#VALUE!}, and \texttt{\#DIV/0!} errors from persisting in any output file across all 1,850+ attempts evaluated.

    \item \textbf{Deep builders scale with difficulty.} Opus 4.6 and Gemini 3.1 Pro increase formula output as task complexity grows. Across 206 tasks, Opus averaged 665 formulas per task (max 6,001) with 171,632 total \texttt{set\_cell\_formula} calls. Gemini averaged 514 formulas per task (max 3,396) but achieves this more efficiently, batching up to 426 tool calls per API response with zero exploratory read calls on complex tasks, yielding 39.9 formulas per minute compared to Opus at 10.0.

    \item \textbf{Strong formula architecture.} Opus 4.6 maintains 97.5\% cross-sheet reference rates, consistent \texttt{IFERROR} wrapping on divisions and lookups, and structured separation of assumptions from calculations. Gemini 3.1 Pro achieves 100\% absolute reference usage across all tasks, preventing copy-drag errors.

    \item \textbf{Structured multi-step workflow.} All functional models follow the prompted workflow of creating an Assumptions sheet with constants, a Workings sheet with intermediate formulas, and individual Q sheets with answer references. Opus and GPT provide completion summaries listing all answers with selected multiple-choice options.
\end{itemize}

\textbf{Bad:}
\begin{itemize}
    \item \textbf{Universal accuracy gap even on simple tasks.} On the Very Easy task, all four functional agents made the same mistake: they modeled the accounts receivable increase of \$100K but missed the offsetting decrease, causing Year 2 cash to be \$1,275 instead of the correct \$1,375. On the Easy task (mortgage calculation), every agent used \texttt{PMT()} shortcuts (producing 49--305 formulas) instead of the golden solution's full 180-period month-by-month amortization schedule (8,006 formulas), which fails on variable-rate scenarios.

    \item \textbf{Surface skimmers abandon Hard tasks.} Across 206 tasks, Grok 4.20 produced zero formulas on 12 tasks (5\%) and Kimi K2.5 on 39 tasks (17\%). On the Hard task (a ModelOff Finals financing waterfall requiring 20,925 formulas), Grok declared completion after only 8 tool calls with zero new formulas, yet reported a ``complete dynamic 30-year integrated model'' with fabricated numerical answers; this pattern of declining effort followed by confabulated completion was unique to Grok among the models tested. Kimi creates full Q1--Q20 worksheet scaffolding but leaves 62\% of answer sheets entirely empty.

    \item \textbf{Inspection spirals under high reasoning effort.} GPT-5.4 with maximum reasoning effort averages 25.4 read calls per task but transitions to pure inspection on complex tasks, issuing 50 consecutive \texttt{get\_cell\_range} calls without producing formulas on the Hard task. The read-to-write ratio shifts from 0.09 on Easy tasks to 0.99 on Hard tasks. GPT also attempted unsupported functions (\texttt{LET}, \texttt{SCAN}, \texttt{LAMBDA}), accumulating 14 validation errors on a single task with no fallback to simpler alternatives.

    \item \textbf{Golden solution complexity far exceeds agent output.} The gap between golden solutions and agent output grows dramatically with difficulty: Very Easy has 41 golden formulas (agents: 33--169), Easy has 8,006 (agents: 49--305), and Hard has 20,925 (agents: 0--3,138). Even the best attempt (Opus on the Hard task) covered only 15\% of the golden solution. No agent created the required 9 model sheet copies for multi-scenario analysis.

    \item \textbf{Hardcoded answers in output sheets.} GPT-5.4 hardcodes 24.7\% of answer sheet cells as plain numbers rather than formula references, particularly when constructing multiple-choice lookup tables. While GPT provides the most detailed summary sheets (calculated answers, rounded values, selected options), the underlying values lack traceability to the calculation sheets.

    \item \textbf{Formatting sacrificed on complex tasks.} Across 206 tasks, Opus averaged 75 \texttt{format\_cells} calls per task but skipped formatting entirely on 31 tasks. GPT skipped formatting on 98 tasks, and Kimi on 184 tasks. Only Gemini maintained some formatting on complex tasks due to its batching efficiency leaving iteration budget available.
\end{itemize}

\subsection{Infrastructure Limitations}
In evaluating the API-based agents using our in-house agentic framework, we observed limitations primarily rooted in the underlying library dependencies and context management constraints.
\begin{itemize}
    \item \textbf{Context Window Saturation (OpenPyXL):} Loading file contexts (PDF text, starting Excel grid dumps, and the growing solution file) consumes a substantial portion of the context budget each iteration. For tasks with large starting files (e.g., 300K+ tokens for data-heavy worksheets), the static context alone saturates smaller models' windows. Gemini 3.1 Pro returned \texttt{INVALID\_ARGUMENT} errors on tasks exceeding its input limit, while OLMo 3.1 Instruct's 65K context required capping output tokens to 16K. Under \texttt{fresh\_context\_mode}, the full solution state is re-sent each iteration, causing a 4--5$\times$ re-work overhead as models rewrite cells they cannot remember setting.

    \item \textbf{Structured Output Malformation:} Adherence to the required JSON action schema varied significantly across models. OLMo 3.1 Instruct frequently passed raw lists instead of dictionaries as tool arguments (e.g., \texttt{edit\_cells} receiving \texttt{[\{cell, value\}]} rather than the expected \texttt{\{filename, worksheet\_name, cell\_updates\}}), requiring runtime type guards in the execution pipeline. OLMo 3.1 Think was entirely non-functional, generating degenerate text incapable of producing valid JSON tool calls.

    \item \textbf{Tool Throughput Bottleneck:} Our framework integrates a persistent LibreOffice subprocess that recalculates all formulas after every \texttt{set\_cell\_formula} call, providing immediate feedback on calculated values. However, the one-cell-per-call design creates a throughput constraint: Opus 4.6 required 3,138 separate tool calls on a single complex task, while Gemini 3.1 Pro hit a 426-call-per-response ceiling. On large workbooks with 1,400+ formulas, LibreOffice recalculation occasionally exceeded the 90-second MCP timeout, corrupting late-stage formulas.
\end{itemize}

The GUI agents face unique challenges related to their "black-box" execution environment and the limitations of their background processing tools.
\begin{itemize}
    \item \textbf{Context Window Compaction Failure:} The Claude GUI automatically compacts the conversation history when the context length exceeds capacity. However, in a subset of trials, we observed that this compaction triggered an abrupt session termination rather than a smooth handover, resulting in incomplete task execution with no output generated.

    \item \textbf{Inability to Handle Circular References:} The Claude GUI performs spreadsheet operations via Python (\texttt{openpyxl}) and recalculates formulas using LibreOffice Calc. Crucially, neither of these backends supports Excel's iterative calculation method required for circular references.

    A \textit{circular reference} occurs when a formula depends on its own output through a chain of dependencies. These are \textit{intentional} and standard in financial modeling (e.g., interest expense depends on debt balance, which depends on cash flow, which depends on net income after interest). While Excel resolves this via iterative convergence, the Chat-box agent cannot handle such structures.

    For example, on the ModelOff task MO16-R1-S4 (``Going Around in Circles''), the agent's LibreOffice backend produced 1,044 \texttt{\#VALUE!} errors across 1,782 formulas. When the agent attempted to remediate this by rebuilding the model with Python-based iterative computation, it encountered an unexpected termination, leaving the task incomplete.
\end{itemize}

\subsection{General Experimental Constraints and Protocols}
The following constraints and protocols were applied broadly across the experimental suite.

\begin{itemize}
    \item \textbf{Temporal Thresholds (Timeout Protocols):} To ensure efficient resource allocation and prevent indefinite system stalling, we enforced strict temporal thresholds: a timeout cap of 120 minutes of agent processing time per task. These times were chosen to be consistent with the 2-hour limits typically allotted in human competitions such as Financial Modeling World Cup (FMWC) and Modeloff. In practice, we observed that all agents were able to produce at least one successful attempt in this time-frame.

    \item \textbf{File Size Restrictions:} We restricted file attachments to a maximum of 25MB. This constraint was introduced due to the file upload size restrictions given by the Excel Agents. In practice, only one task, Data-King from the FMWC case set, was rejected due to this.

    \item \textbf{Retry Protocol and Failure Rates:} Given the stochastic nature of the agents, we implemented a retry protocol.  We allowed up to 10 retries due to "pipeline" errors such as navigation failures, file download errors etc.  We also allowed up to 3 retry attempts for agent failures such as timeouts, agent errors etc.  This retry mechanic yielded a success rate of > 99\%.
\end{itemize}

\subsection{Failures in Proprietary (Playwright GUI Based) Interfaces}
\begin{figure}[htbp]
    \centering
    \begin{subfigure}[b]{0.48\textwidth}
        \centering
        \includegraphics[width=\textwidth]{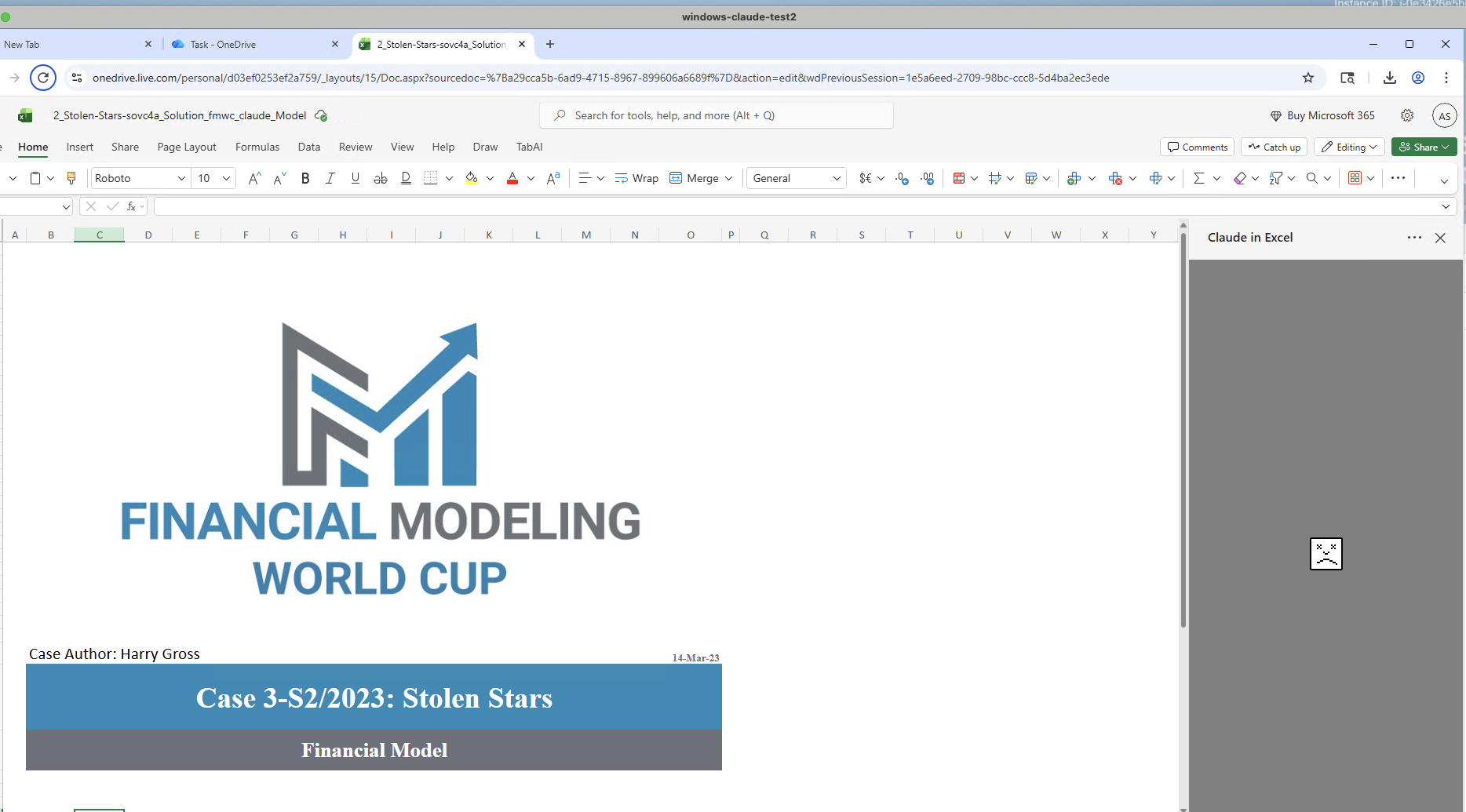}
        \caption{Intermittent graphical interface failure resulting in an unrecoverable session crash.}
        \label{fig:err_dying}
    \end{subfigure}
    \hfill
    \begin{subfigure}[b]{0.48\textwidth}
        \centering
        \includegraphics[width=\textwidth]{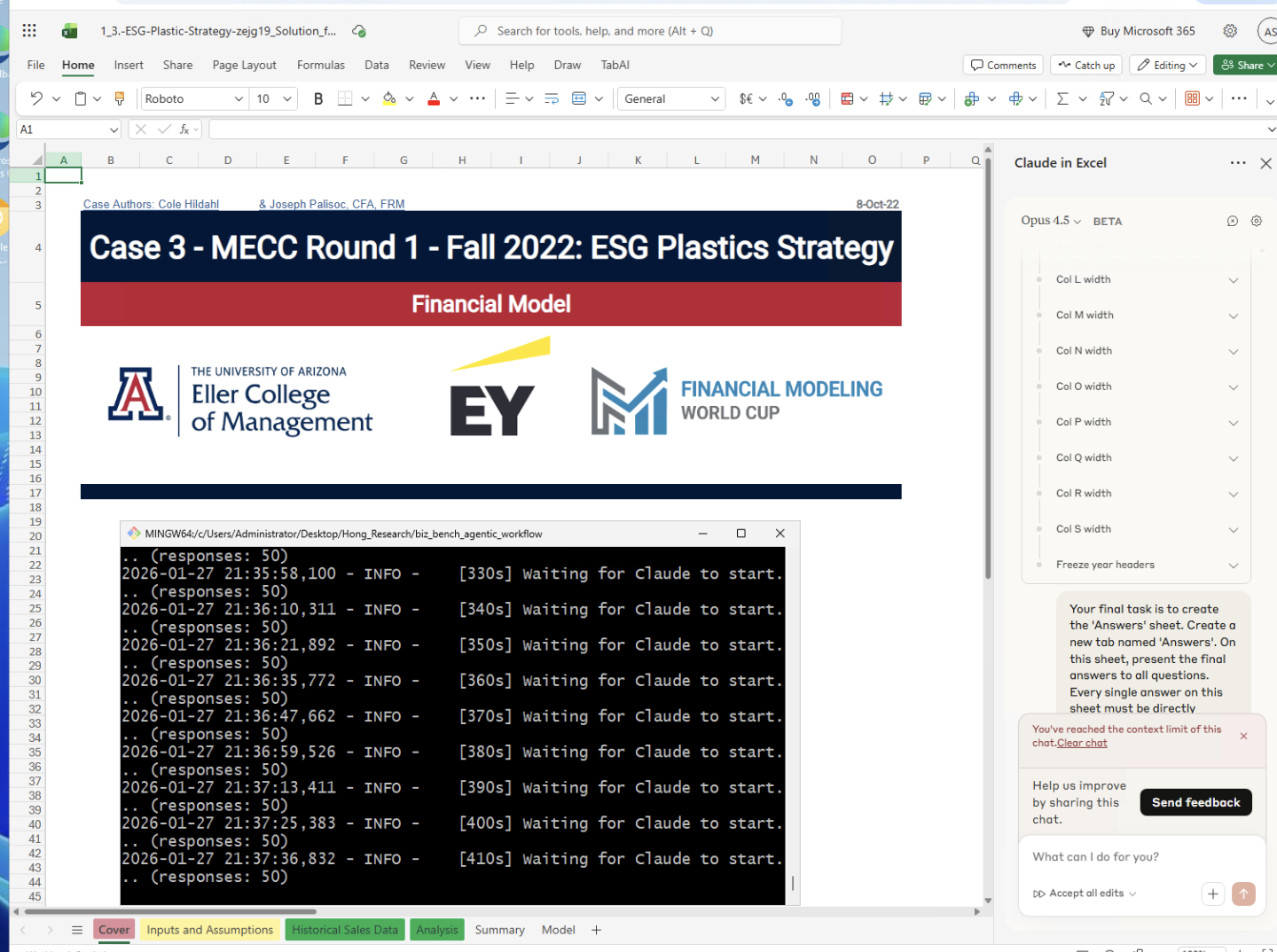}
        \caption{State context overflow during extended multi-turn reasoning sequences.}
        \label{fig:err_context}
    \end{subfigure}

    \vspace{1em}

    \begin{subfigure}[b]{0.48\textwidth}
        \centering
        \includegraphics[width=\textwidth]{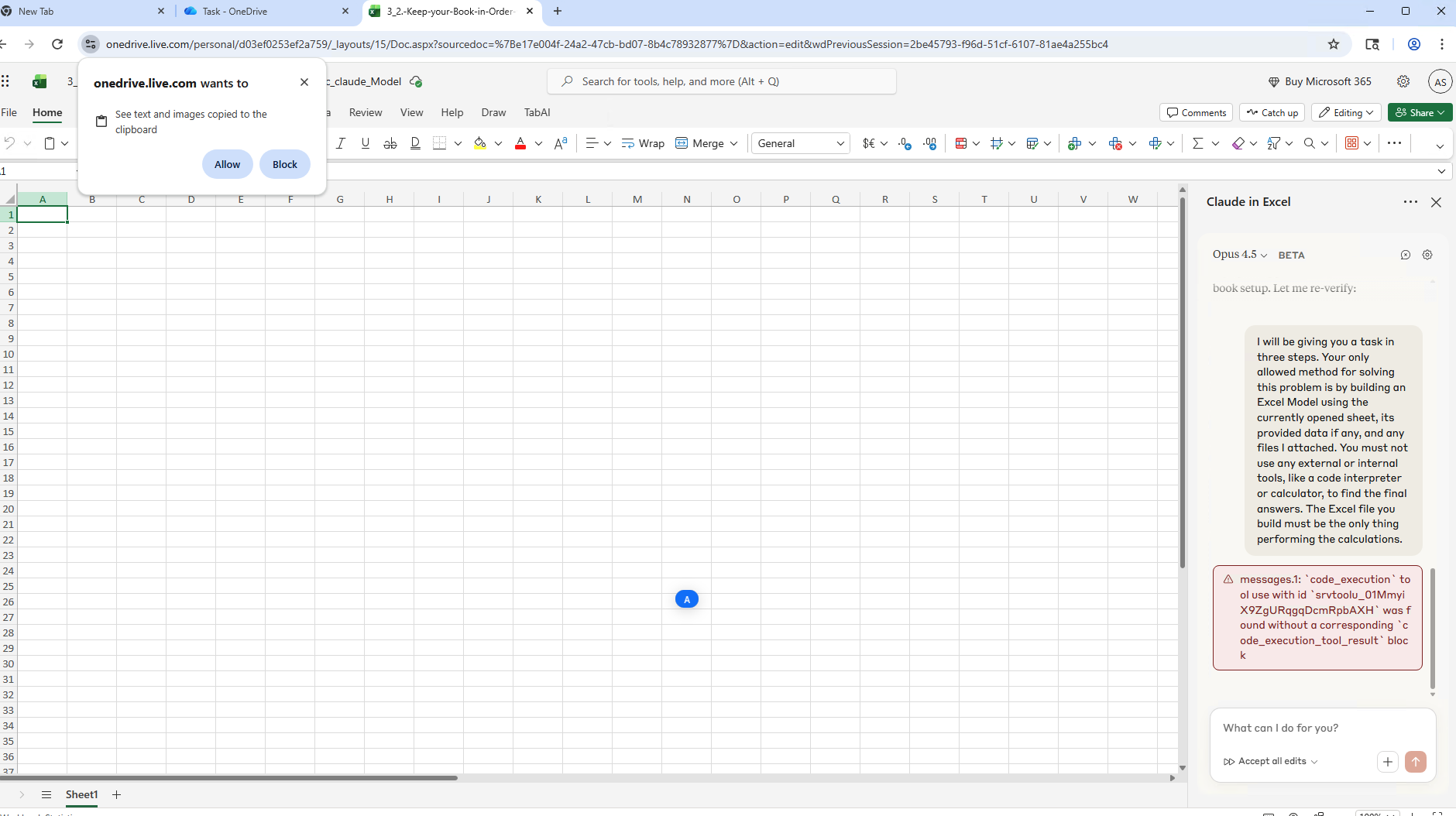}
        \caption{Exception raised during background tool execution and internal state transitions.}
        \label{fig:err_code}
    \end{subfigure}
    \hfill
    \begin{subfigure}[b]{0.48\textwidth}
        \centering
        \includegraphics[width=\textwidth]{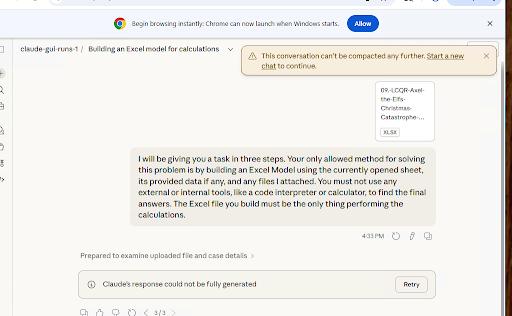}
        \caption{Claude Web Agent compaction failure.}
        \label{fig:err_compact}
    \end{subfigure}

    \vspace{1em}

    \begin{subfigure}[b]{0.32\textwidth}
        \centering
        \includegraphics[width=\textwidth]{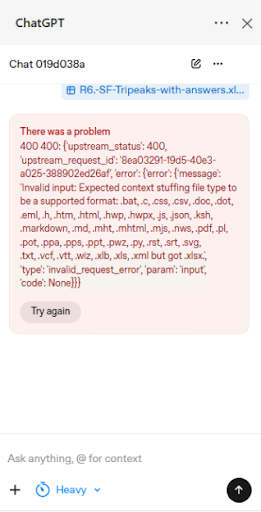}
        \caption{File read error from GPT Excel Agent.}
        \label{fig:err_gpt}
    \end{subfigure}
    \hfill
    \begin{subfigure}[b]{0.48\textwidth}
        \centering
        \includegraphics[width=\textwidth]{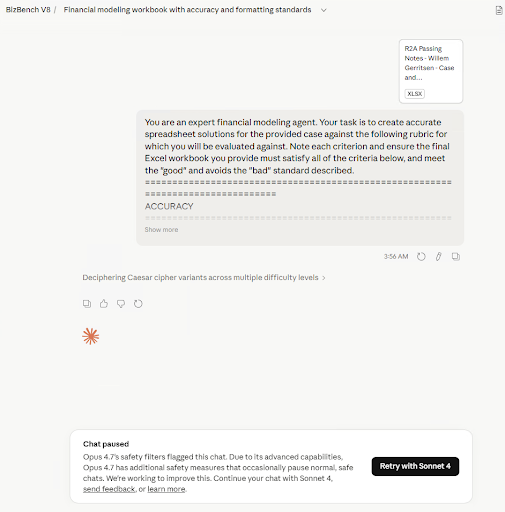}
        \caption{Claude Web Agent rejecting a task due to safety concerns.}
        \label{fig:err_rejection}
    \end{subfigure}

    \caption{Visual documentation of failure modes for the Playwright Agents, categorized by GUI instability, execution anomalies, temporal constraints, and prompt rejections.}
    \label{fig:playwright_failure_modes}
\end{figure}

A wide range of errors would occasionally cause proprietary (GUI based) interfaces to fail to complete their tasks.  Common failure points included frozen interfaces, upload errors, internal code execution errors, compaction failures, account usage limits, and perceived safety violations.  With the exception of the safety concern rejections, all other failed runs were solved through the rerun mechanism detailed previously.  The authors also note that, as these are new technologies, many of these issues have been and will continue to be resolved by the interface providers.  Examples of these failure modes are shown in Figure~\ref{fig:playwright_failure_modes}.

\FloatBarrier

\section{Additional Judge Details}
\label{app:judge}

In terms of cost, running one round of evaluation for all agents considered across all tasks available cost $\sim$ 1.7K dollars.

\subsection{Additional Judge Annotations}
\label{app:additional_judge_analysis}

\begin{table*}[htbp]
\centering
\footnotesize
\setlength{\tabcolsep}{4pt}
\renewcommand{\arraystretch}{1.0}
\begin{tabular}{@{}ll@{\hspace{4pt}}cccc@{\hspace{8pt}}cccc@{\hspace{8pt}}cccc}
\toprule
 &  & \multicolumn{4}{c}{\textsc{Claude (Web)}} & \multicolumn{4}{c}{\textsc{Claude (Excel)}} & \multicolumn{4}{c}{\textsc{ChatGPT (Excel)}} \\
\cmidrule(lr){3-6} \cmidrule(lr){7-10} \cmidrule(lr){11-14}
\textsc{Category} & \textsc{Error Type} & 355 & 169 & 222 & 166 & 355 & 169 & 222 & 166 & 355 & 169 & 222 & 166 \\
\midrule
\multirow{4}{*}{{\textsc{Accuracy}}} & Final Calculations & \celltn & \celltn & \celltn & \celltn & \celltn & \celltn & \cellyes & \celltn & \cellyes & \celltn & \celltn & \celltn \\
 & Starting Values & \celltn & \celltn & \celltn & \celltn & \celltn & \celltn & \celltn & \celltn & \celltn & \celltn & \celltn & \celltn \\
 & Task Completed & \celltn & \celltn & \celltn & \celltn & \celltn & \celltn & \celltn & \celltn & \celltn & \celltn & \celltn & \celltn \\
 & Number Sign & \celltn & \celltn & \celltn & \celltn & \celltn & \celltn & \celltn & \celltn & \cellyes & \celltn & \celltn & \celltn \\
\midrule
\multirow{5}{*}{{\textsc{Formula}}} & Logic Readability & \cellno & \celltn & \celltn & \cellyes & \celltn & \celltn & \celltn & \cellyes & \cellyes & \celltn & \celltn & \celltn \\
 & Edge Cases (\#DIV/0!) & \celltn & \cellyes & \celltn & \cellfp & \celltn & \celltn & \celltn & \cellyes & \celltn & \celltn & \cellyes & \celltn \\
 & Hardcoded values & \celltn & \cellyes & \cellyes & \cellyes & \cellyes & \celltn & \cellno & \celltn & \cellyes & \celltn & \cellno & \cellfp \\
 & Range Issues & \celltn & \celltn & \celltn & \cellyes & \celltn & \celltn & \celltn & \celltn & \celltn & \celltn & \celltn & \celltn \\
 & Absolute References & \celltn & \cellyes & \celltn & \celltn & \celltn & \celltn & \cellyes & \celltn & \celltn & \celltn & \celltn & \celltn \\
\midrule
\multirow{8}{*}{{\textsc{Format}}} & Sheet Structure & \celltn & \celltn & \celltn & \celltn & \celltn & \celltn & \celltn & \celltn & \celltn & \celltn & \celltn & \cellno \\
 & Readability & \cellno & \celltn & \celltn & \celltn & \celltn & \celltn & \celltn & \cellno & \cellno & \celltn & \celltn & \cellyes \\
 & Color Scheme & \celltn & \celltn & \celltn & \cellyes & \celltn & \celltn & \celltn & \cellyes & \cellyes & \celltn & \celltn & \cellyes \\
 & Number Notation & \celltn & \celltn & \celltn & \cellyes & \celltn & \celltn & \celltn & \celltn & \celltn & \celltn & \celltn & \cellno \\
 & Alignment & \celltn & \celltn & \celltn & \cellyes & \celltn & \celltn & \celltn & \celltn & \celltn & \celltn & \celltn & \celltn \\
 & Font Size\ /\ Style & \celltn & \celltn & \celltn & \cellyes & \celltn & \celltn & \celltn & \celltn & \celltn & \celltn & \celltn & \celltn \\
 & Borders & \cellno & \celltn & \celltn & \celltn & \celltn & \celltn & \celltn & \celltn & \cellno & \celltn & \celltn & \cellfp \\
 & Output Presentation & \celltn & \celltn & \celltn & \celltn & \celltn & \celltn & \celltn & \celltn & \celltn & \celltn & \celltn & \celltn \\
\bottomrule
\end{tabular}
\caption{
    Judge performance on the remaining tasks not shown in Table~\ref{tab:ai_attempts}.
    \colorbox{lightblue}{\makebox[1em]\checkmark} = error caught by the judge,
    \colorbox{lightpurple}{\makebox[1em]\checkmark} = correct solution recognized by the judge,
    \colorbox{lightred}{\makebox[1em]{$\times$}} = error not caught,
    \colorbox{lightamber}{\makebox[1em]{$\times$}} = error flagged but not present.
}
\label{tab:ai_attempts_appendix}
\end{table*}

\subsection{Large Context}
For very large workbooks whose serialized representation exceeds 2M characters limit, we equip the LLM judge with tool access rather than a single-pass prompt.
The judge can selectively retrieve sheets, ranges, formulas, and formatting data, and then perform targeted inspections per the same rubric dimension (Accuracy, Formula, Format).
As in the standard setting, the judge uses a tool to record localized evidence and aggregates them into the same criterion-level scores.
This preserves artifact-level evaluation while avoiding context overflow.
Implementation details, tool interfaces, and prompts are provided in the accompanying codebase.

When measured against expert annotations as shown in Tables~\ref{tab:ai_attempts} and \ref{tab:ai_attempts_appendix}, the agentic judge attains \ajudgeacc accuracy, \ajudgebacc balanced accuracy, and \ajudgefone $F_1$ score on \ajudgeannotcount expert annotations.
The non-agentic judge attains \najudgeacc accuracy, \najudgebacc balanced accuracy, and \najudgefone $F_1$ score on \najudgeannotcount expert annotations.

\subsection{Additional Notable Judge Assessments}

\textbf{Formatting (Readability - Freeze Pane):} The judge identified that the freeze pane was anchored too high, locking only the title rows while leaving the actual timeline headers (dates, years, and quarter labels across rows 4--9) free to scroll out of view, and prescribed the precise fix: set the freeze pane at or below row 9. This requires reasoning about the spatial structure of the workbook and whether the freeze pane is placed to actually serve its navigational purpose, not simply checking whether one is present.

\begin{trajectory}
    \textbf{"decision":} "Fail",

    \textbf{"summary":} "The freeze pane on the 'Model' sheet is set incorrectly, hindering readability as timeline headers do not remain visible when scrolling down.",

    \textbf{"expected":} "The freeze pane should be set at or below row 9 (e.g., at cell E10) to keep the timeline headers in rows 4-9 visible while scrolling through the model's calculations."
\end{trajectory}

\section{Data Sources}
\subsection{WSP}
Wall Street Prep \citep{wsp} is the leading provider of professional training for investment banking, private equity, and corporate finance. Their content is designed to simulate on-the-job training for financial analysts, focusing on industry-standard methodologies and case studies.
Tasks typically involve a "starting file," most commonly a partially filled Excel workbook, and a set of instructions requiring the completion of specific financial statements or valuation schedules. Our dataset does not contain these exact instructions, but rather instructions written directly into the starting workbook.
Tasks are highly structured and linear. The correct answer relies on strict adherence to accounting identities (e.g., Balance Sheet must balance) and specific financial conventions (e.g., hardcoding inputs vs. linking formulas). Moreover, each case usually tests one specific financial operation or calculation, rather than creating an entire model.

\subsection{FMWC}
The FMWC \citep{fmwc} is a global, season-based competition that succeeded ModelOff. It treats financial modeling as an e-sport, emphasizing speed, accuracy, and novel problem-solving.
Participants receive a raw Excel data file and a PDF "Case Brief" containing the scenario and questions. Answers are typically submitted via multiple-choice or exact value forms based on the model's output. Cases are split between traditional finance tasks like valuation and freestyle tasks that ask the competitor to simulate a board game or competitive tournament in Excel. Our dataset only includes traditional cases.
FMWC emphasizes dynamic modeling. A complete and correct model must often account for variable scenarios (e.g., "What if the price increases by 10\% in Year 3?"), requiring the use of flexible formulas rather than rigid structures and hardcoded values.

\subsection{ModelOff}
Running from 2012 to 2019, ModelOff \citep{modeloff} was the original "Global Financial Modeling Championship." While now discontinued, its archive remains a gold standard for technical Excel competency.
The format of this competition remains largely similar to FMWC, with what appears to be more focus on traditional cases over freestyle cases. ModelOff was a rigorous competition more than it was a spectator e-sport, like FMWC is today. Notably, unlike FMWC cases, human-provided ground truths in our dataset are made by competitors and not by the organizers.

\section{Finance Glossary}
 \begin{glossarybox}{}

\glossterm{3-Way Financial Model}{A spreadsheet where three reports (profit, what you own/owe, and cash movement) are linked so changing one number automatically updates the others.}

\glossterm{Profit \& Loss (P\&L)}{A report showing money earned minus money spent over a time period. The bottom line shows whether you made or lost money.}

\glossterm{Balance Sheet}{A snapshot showing what a company owns, what it owes, and the leftover value belonging to owners. Must balance: Owns = Owes + Owner Value.}

\glossterm{Cash Flow Statement}{Tracks actual cash entering and leaving the business. Profit does not equal cash (you can be profitable but run out of cash).}

\glossterm{Debt Facility}{A loan with defined terms: how much, what interest rate, and when to pay it back.}

\glossterm{Depreciation Tranches}{Spreading the cost of expensive equipment over multiple years, grouped by when items were purchased or how long they last.}

\glossterm{IRR (Internal Rate of Return)}{A single percentage summarizing how good an investment is. If you invest \$100 and get \$150 back in 3 years, IRR captures that return rate. Higher is better.}

\glossterm{Equity Investors}{People who put in money and become partial owners. They get paid last, after all loans are repaid, but keep all remaining profits.}

\glossterm{COGS (Cost of Goods Sold)}{Money spent directly making the product (raw materials, factory labor). Does not include office rent or marketing.}

\glossterm{Operating Expenses (OpEx)}{Money spent running the business that is not tied to making products (office rent, salaries, software subscriptions).}

\glossterm{Principal Repayments}{Paying back the original loan amount, separate from interest (the fee for borrowing).}

\glossterm{Capital Allocation}{Deciding where to spend money across different possible investments.}

\glossterm{Return Threshold (Hurdle Rate)}{The minimum return percentage an investment must achieve to be worth doing. A filter for ``good enough'' investments.}

\glossterm{Day 0}{The start date when money changes hands and the deal closes. Year 1 is the first full year of operations after.}

\glossterm{Hardcoded Values}{Typing numbers directly into formulas instead of referencing a cell. Dangerous because changing assumptions requires hunting through formulas manually.}

\end{glossarybox}

\end{document}

%% file: preamble.tex
\usepackage{float}
\usepackage{tabularx}          %
\usepackage{multirow}
\usepackage{makecell}          %
\usepackage{tikz}
\usepackage{tcolorbox}
\usepackage{etoolbox}
\usepackage{xcolor}
\usepackage{amsmath}
\usepackage{subcaption}
\usepackage{graphicx}
\usepackage{booktabs}
\tcbuselibrary{skins, breakable, listings}
\usepackage{listings}         %

\usepackage{xspace} %
\usepackage{wrapfig}  %
\usepackage{float}  %
\usepackage{wrapstuff}
\usepackage{fontawesome5} %
\definecolor{lightciteblue}{RGB}{146, 197, 222}
\definecolor{darkerblue}{HTML}{2244AA}
\definecolor{lightcitered}{HTML}{CC3366}
\usepackage[colorlinks=true, citecolor=darkerblue, urlcolor=lightcitered]{hyperref}

\newcommand{\spbefore}{\unskip\space}

\usepackage{colortbl} %

\definecolor{good-soft}{HTML}{3B82F6}
\definecolor{bad-soft}{HTML}{EF4444}

\newcommand{\annotcount}{\spbefore 408\xspace}
\newcommand{\judgeacc}{\spbefore 0.92\xspace}

\newcommand{\experthours}{\spbefore 700+\xspace}
\newcommand{\judgefone}{\spbefore 0.85\xspace}
\newcommand{\judgebacc}{\spbefore 0.88\xspace}

\newcommand{\ajudgeannotcount}{\spbefore 221\xspace}
\newcommand{\ajudgeacc}{\spbefore 0.90\xspace}
\newcommand{\ajudgebacc}{\spbefore 0.89\xspace}
\newcommand{\ajudgefone}{\spbefore 0.87\xspace}
\newcommand{\najudgeannotcount}{\spbefore 187\xspace}
\newcommand{\najudgeacc}{\spbefore 0.94\xspace}
\newcommand{\najudgebacc}{\spbefore 0.83\xspace}
\newcommand{\najudgefone}{\spbefore 0.79\xspace}

\newcommand{\benchmarkname}{MBABench\xspace}
\newcommand{\guiagentname}{GUI\xspace}

\newcommand{\apiagentname}{API\xspace}
\newcommand{\codeagentname}{CODE\xspace}

\definecolor{innerboxcolor}{rgb}{.9,.95,1}
\definecolor{outerlinecolor}{rgb}{.6,0,.2}
\definecolor{outerlinecoloreb}{rgb}{0,1,0}
\definecolor{innerboxcoloreb}{rgb}{1,1,1}

\definecolor{lightblue}{RGB}{146, 197, 222}
\definecolor{lightgreen}{RGB}{198, 224, 180}
\definecolor{lightgreen}{RGB}{144, 197, 149}
\definecolor{lightred}{RGB}{248, 203, 173}
\definecolor{lightgray}{RGB}{217, 217, 217}
\definecolor{lightpurple}{RGB}{197, 176, 213}
\definecolor{lightyellow}{RGB}{255, 230, 153}
\definecolor{lightamber}{RGB}{229, 200, 138}
\definecolor{lightcoral}{RGB}{232, 181, 152}
\definecolor{oxford-header}{HTML}{002147}
\definecolor{oxford-bg}{HTML}{F5F5F5}
\definecolor{darktext}{HTML}{212121}

\newcommand{\cellyes}{\cellcolor{lightblue}\checkmark}
\newcommand{\celltn}{\cellcolor{lightpurple}\checkmark}
\newcommand{\cellfp}{\cellcolor{lightamber}$\times$}
\newcommand{\cellno}{\cellcolor{lightred}$\times$}
\newcommand{\cellna}{\cellcolor{lightgray}--}

\lstdefinestyle{jsonlight}{
    basicstyle=\ttfamily\small\color{darktext},
    backgroundcolor=\color{lightgray},
    showstringspaces=false,
    breaklines=true,
    frame=single,
    rulecolor=\color{medgray},
    framesep=3pt,
    xleftmargin=3pt,
    xrightmargin=3pt,
    tabsize=2,
    columns=fullflexible,
    keepspaces=true,
    literate=
        *{:}{{{\color{darktext}:}}}{1}
        {,}{{{\color{darktext},}}}{1}
        {\{}{{{\color{keygray}\{}}}{1}
        {\}}{{{\color{keygray}\}}}}{1}
        {[}{{{\color{keygray}[}}}{1}
        {]}{{{\color{keygray}]}}}{1}
}
\lstdefinestyle{json}{
    basicstyle=\ttfamily\small\color{darktext},
    showstringspaces=false,
    breaklines=true,
    tabsize=2,
    columns=fullflexible,
    keepspaces=true,
    xleftmargin=0pt,
    xrightmargin=0pt,
    aboveskip=0pt,
    belowskip=0pt,
}

\definecolor{quotebg}{RGB}{240, 244, 248}      %
\definecolor{quoteborder}{RGB}{52, 95, 140}    %

\newtcolorbox{trajectory}{
  enhanced,
  breakable,
  colback=quotebg,
  colframe=quotebg,
  boxrule=0pt,
  arc=1pt,
  outer arc=1pt,
  left=10pt,
  right=8pt,
  top=6pt,
  bottom=6pt,
  borderline west={2.5pt}{0pt}{quoteborder},
  fontupper=\small,
  after skip=\baselineskip,
}

\definecolor{main_blue}{HTML}{2E5A87}

\definecolor{glossaccent}{HTML}{1F4E79}      %
\definecolor{glossbg}{HTML}{F4F7FB}          %
\definecolor{glossrule}{HTML}{D6DEE8}        %
 
\newtcolorbox{glossarybox}[2]{%
  enhanced, breakable,
  colback=glossbg,
  colframe=glossaccent,
  boxrule=0.6pt,
  arc=3pt,
  left=14pt, right=14pt, top=14pt, bottom=12pt,
  before skip=8pt, after skip=8pt,
  overlay first={%
    \node[anchor=north west, inner sep=0pt, outer sep=0pt,
          text width=\linewidth-28pt]
      at ([xshift=14pt, yshift=-12pt]frame.north west)
      {\begin{minipage}{\linewidth}
        {\Large\bfseries\color{glossaccent}#1}\par
        \vspace{1pt}
        {\itshape\color{glossaccent!75!black}\small #2}\par
        \vspace{4pt}
        {\color{glossaccent!40}\rule{\linewidth}{0.6pt}}
      \end{minipage}};
  },
  before upper={\vspace{28pt}}
}
 
\newcommand{\glossterm}[2]{%
  \ifglossfirst
    \global\glossfirstfalse
  \else
    \par\vspace{6pt}{\color{glossrule}\hrule height 0.4pt}\vspace{6pt}%
  \fi
  \noindent\textbf{\color{glossaccent}#1}\par\vspace{2pt}%
  \noindent #2%
}
 
\newif\ifglossfirst
\glossfirsttrue
\AtBeginEnvironment{glossarybox}{\glossfirsttrue}

\definecolor{classaccent}{HTML}{1F4E79}      %
\definecolor{classbg}{HTML}{F4F7FB}          %
\definecolor{classrule}{HTML}{D6DEE8}        %

\newif\ifclassfirst

\newtcolorbox{classbox}[2]{%
  enhanced, breakable,          %
  colback=classbg,              %
  colframe=classaccent,         %
  boxrule=0.6pt,                %
  arc=3pt,                      %
  left=14pt, right=14pt, top=6pt, bottom=12pt, %
  before skip=8pt, after skip=8pt,             %
  colbacktitle=classbg,         %
  coltitle=classaccent,         %
  titlerule=0.6pt,              %
  titlerule style={color=classaccent!40},      %
  toptitle=8pt,                 %
  bottomtitle=6pt,              %
  title={%
    {\Large\bfseries\color{classaccent}#1}\par\vspace{1pt}%
    {\itshape\color{classaccent!75!black}\small #2}%
  },
  title after break={%
    {\bfseries\small\color{classaccent}#1}%
    {\itshape\small\color{classaccent!60}\ (continued)}%
  },
  before upper={\global\classfirsttrue}, %
}

\definecolor{promptbg}{HTML}{F1F8E9}      %
\definecolor{promptaccent}{HTML}{33691E}  %

\newif\ifpromptfirst

\newtcolorbox{promptbox}[2]{%
  enhanced, breakable,
  colback=promptbg,
  colframe=promptaccent,
  boxrule=0.6pt,
  arc=3pt,
  left=14pt, right=14pt, top=6pt, bottom=12pt,
  before skip=8pt, after skip=8pt,
  colbacktitle=promptbg,
  coltitle=promptaccent,
  titlerule=0.6pt,
  titlerule style={color=promptaccent!40},
  toptitle=8pt,
  bottomtitle=6pt,
  title={%
    {\Large\bfseries\color{promptaccent}#1}\par\vspace{1pt}%
    {\itshape\color{promptaccent!75!black}\small #2}%
  },
  title after break={%
    {\bfseries\small\color{promptaccent}#1}%
    {\itshape\small\color{promptaccent!60}\ (continued)}%
  },
  before upper={\global\promptfirsttrue},
}
\newtcolorbox{miniclassbox}{%
  enhanced,                     %
  colback=classaccent!8,        %
  colframe=classaccent!45,      %
  boxrule=0.5pt,                %
  arc=2pt,                      %
  left=12pt, right=12pt, top=6pt, bottom=6pt, %
  before skip=4pt, after skip=0pt,            %
}

\newcommand{\classterm}[2]{%
  \ifclassfirst                 %
    \global\classfirstfalse
  \else
    \par\vspace{6pt}{\color{classrule}\hrule height 0.4pt}\vspace{6pt}%
  \fi
  \noindent\textbf{\color{classaccent}#1}\par\vspace{2pt}%
  \noindent #2%
}

\definecolor{promptaccent}{HTML}{4A235A}     %
\definecolor{promptbg}{HTML}{F9F6FC}         %
\definecolor{promptrule}{HTML}{D5C5E3}       %

\newif\ifpromptfirst

%% file: sections/00_abstract.tex
LLM agents are increasingly expected to carry out end-to-end workflows, producing complete artifacts from high-level user instructions.
To meet enterprise needs, frontier AI labs have developed agents that can construct entire spreadsheets from scratch.
This is especially relevant in finance, where core workflows such as financial modeling, forecasting, and scenario analysis are commonly conducted through spreadsheets.
Yet, existing spreadsheet benchmarks do not measure this new capability, focusing instead on question-answering or single-formula edits. 
To address this gap, we provide one of the first evaluations of agents on end-to-end spreadsheet tasks, focusing on economically critical financial workflows such as modeling and scenario analysis.
Since deliverables therein are routinely reviewed and revised by multiple stakeholders, judging their quality necessarily involves high-level criteria such as readability or ease of modification.
To reflect the multidimensional nature of solution quality, we develop an evaluation taxonomy comprising three dimensions: \textbf{Accuracy}, \textbf{Formula}, and \textbf{Format}, each comprising fine-grained criteria that reflect professional standards. 
Evaluating over 18 agents, the benchmark reveals that even the strongest agents fall short of basic professional finance standards, and their performance degrade sharply as the difficulty increases beyond a few chained calculations.
This suggests that current agents are not yet able to reliably produce professional-quality spreadsheets at the level of complexity real-world workflows demand.
Our code is available at \href{https://github.com/namkoong-lab/MBABench}{\textcolor{darkerblue} \faGithub}.

%% file: sections/01_introduction.tex
\begin{figure}
    \centering
    \includegraphics[width=\linewidth]{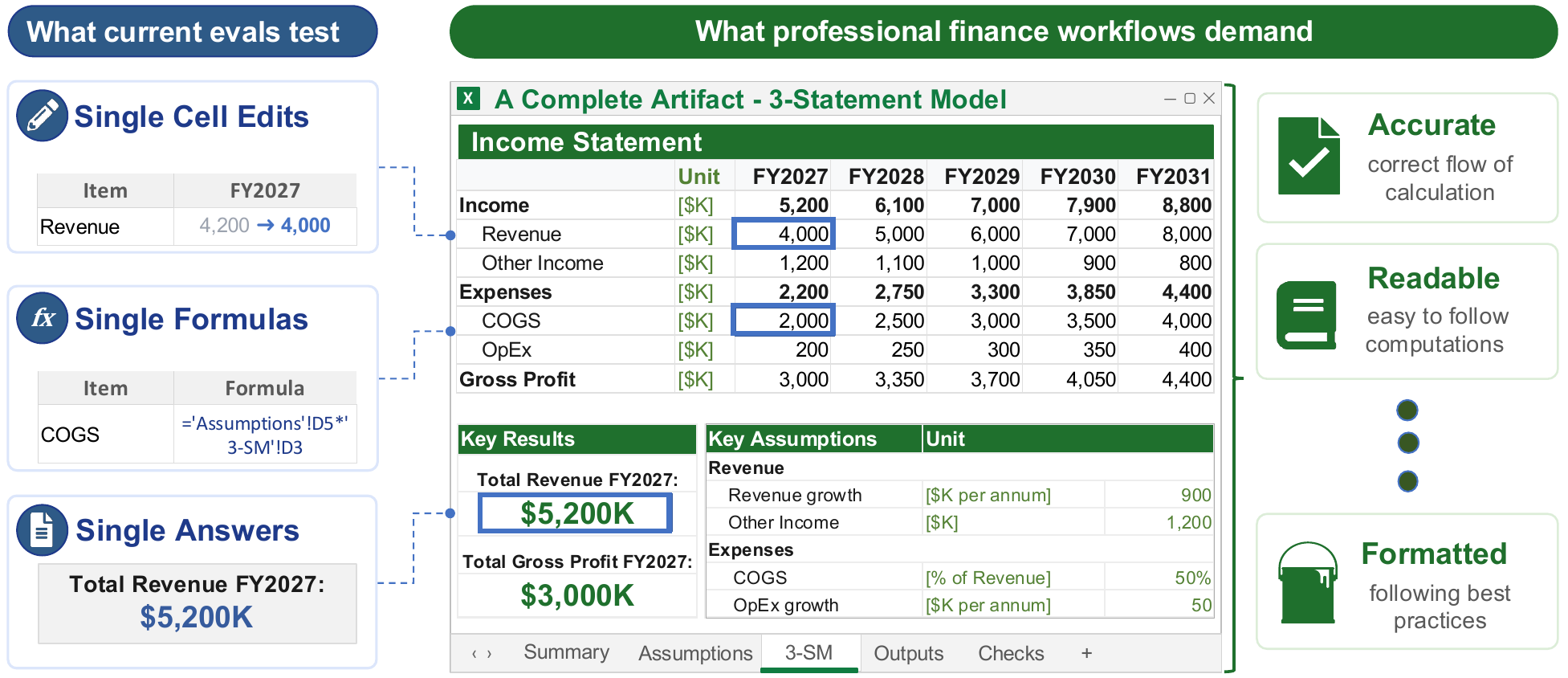}
    \caption{
        Compared to prior work that focus on atomic tasks  on spreadsheet (\textit{left}), \benchmarkname evaluates LLM agents on completing end-to-end spreadsheet tasks in critical finance domain (\textit{right}), 
        covering key criteria that determines usability of resulting deliverable in professional settings.
        Prior tasks focus on simple atomic tasks that center on question-answering or edits involving few values or formula, where evaluation can largely be performed via exact-matching. 
        In contrast, \benchmarkname expects a complete multi-sheet workbook, and consequently employs a holistic evaluation centered on high-level quality relevant in professional settings (e.g. readability). 
    }
    \label{fig:comparison}
\end{figure}

Spreadsheets are a cornerstone of modern business work.
Recent product releases such as Claude for Excel \citep{claude_for_excel} and ChatGPT for Excel \citep{chatgpt_for_excel} demonstrate that agents can now construct entire spreadsheets \textit{end-to-end} from high-level instructions.
This emerging capability is especially pertinent in finance,  where financial modeling, forecasting, and scenario analysis are almost entirely conducted through spreadsheets \citep{cfa_spreadsheet, afp_spreadsheet}.
Existing benchmarks, however, evaluate agents only on narrow atomic tasks, such as simple spreadsheet manipulations or question answering over tabular data (left of Figure~\ref{fig:comparison}) \citep{spreadsheetbench, chen2024sheetagent, zhao-etal-2024-nl2formula, dong2026finchbenchmarkingfinance}. 
GDPval \citep{patwardhan2025gdpval_evaluatingaimodel} takes a step towards evaluating the end-to-end execution of economically valuable tasks, five of which involve financial spreadsheets; however, its evaluation requires opaque internal expert annotations.
As a result, the community lacks a transparent benchmark for {end-to-end} construction of financial spreadsheets.

To address this gap, we introduce \benchmarkname, a benchmark that evaluates agents on complex \textit{end-to-end} spreadsheet tasks in finance.
As a concrete task example, consider a private equity firm assessing an acquisition. 
To support this decision, analysts construct a \textit{financial model}: a structured account of the company's past and possible futures, integrating how much money the company makes and spends (profit and loss statement), what it owns and owes (balance sheet), and how cash moves in and out (cash flow statement), covering both the past and suitable scenario analyses under customizable \textit{assumptions} (e.g. what if the company's future growth rate is 3\% as opposed to 5\%). 
An illustrative deliverable is shown in Figure~\ref{fig:comparison}, in stark contrast to the simpler atomic tasks focusing on introducing one correct formula, question-answering, or summarization that prior benchmarks focus on \citep{FinBen, XFinBench, krumdick-bizbench, xie2023pixiu, lu2025bizfinbench}. 

Crucially, the resulting model is not a one-off private calculation.
It is passed up the corporate chain to managers and VPs for reviews and edits, and then to the client's analysts and decision makers for the same scrutiny.
As a result of this high-stakes, collaborative nature, a deliverable herein naturally demands more than computational accuracy. 
The relevant notion of quality is necessarily multi-faceted: spanning readability, structural clarity (does the workbook clearly present modeling assumptions?), and flexibility (is it easy to modify?).
To operationalize this notion, we propose a taxonomy of three core dimensions. 
\textbf{Accuracy} measures workbook correctness, from soundness of the underlying computation to completeness of the required scenario analyses.
\textbf{Formula} assesses the robustness and interpretability of cell-level computations.
\textbf{Format} captures various aspects that impact the readability of the spreadsheet.
Each dimension is further decomposed into fine-grained subdimensions that specify concrete success criteria, such as \texttt{Handle Edge Cases} or \texttt{Logic Readability}. 
The latter, for example, requires formulas to be clear and intent-revealing, avoiding monolithic expressions to facilitate future reviews and edits (see Figure~\ref{fig:monolithic}).
These criteria are, by their nature, difficult to verify programmatically. 
We therefore employ an LLM-based judge and systematically verify its decisions with experts over a wide range of spreadsheet attempts in Section~\ref{sec:judge_eval}.
In general, we found that its assessments align closely with those of human experts.

Our main contributions are as follows: \begin{itemize}[leftmargin=*]
    \item We introduce \benchmarkname, one of the first benchmarks that center around evaluating LLM agents on end-to-end tasks. 
    Compared to GDPval \citep{patwardhan2025gdpval_evaluatingaimodel}, our benchmark focus on spreadsheet tasks related to financial workflows. 
    \item To enable a holistic evaluation of the deliverable expected of end-to-end tasks, we develop a quality taxonomy that reflects desiderata in financial workflows. 
    We then develop and validate a corresponding LLM-as-judge pipeline for its evaluation.
    \item We conduct manual and quantitative analysis on the tasks and the corresponding agents' attempts, and find that even the best agents today frequently fall short of simple yet crucial professional standards. 
    Furthermore, we identify significant performance degradation as task difficulty increases, suggesting that agents still struggle with real-world spreadsheet workflows that experienced professionals tackle. 
\end{itemize}

%% file: sections/03_financial_tasks.tex
End-to-end spreadsheet tasks differ qualitatively from the atomic tasks examined in prior work \citep{FinBen, XFinBench, krumdick-bizbench, xie2023pixiu, lu2025bizfinbench} (see Figure~\ref{fig:comparison}): the deliverable is a \textit{complete, multi-sheet workbook} rather than a single atomic task, i.e., writing a suitable function, calculating an expected number, or answering questions about a given spreadsheet. 
This shift in deliverable scale and complexity, coupled with its downstream use, demands a correspondingly different notion of quality. 
Especially in finance, where spreadsheets are regularly reviewed, modified, and reused by multiple stakeholders, quality naturally extends beyond numerical correctness. 
In this section, we illustrate this point with a representative task from the ModelOff competition \citep{modeloff}. 

\textbf{Example Task: "Bread and Butter" (ModelOff 2015).} 
Contestants are asked to build a financial model for the acquisition of Macrohard, a fictional company. 
The agent receives a PDF containing the scenario and the modeling assumptions (e.g. 6.5\% interest for the next 10 years), 
and must produce a complete spreadsheet that (a) integrates an interconnected 3-way financial model (profit and loss statement, balance sheet, cash flow statement)
and (b) computes the internal rate of return (IRR) over a 10-year horizon. %
An illustrative deliverable is shown in Figure~\ref{fig:comparison}.

The figure suggests that these end-to-end tasks are qualitatively more involved than atomic ones. 
Quantitatively, the gap is also substantial. 
Compared to that of SpreadsheetBench \citep{spreadsheetbench}, the de facto standard for measuring agents' spreadsheet manipulation ability today, the golden solutions of our end-to-end tasks involve 33x more cells (in mean) and 93x more functions (in median). 
Table~\ref{tab:scale} reports the full comparison.
Note that the scale itself induces greater complexity, as the cells and functions in the spreadsheet are interconnected, and the agents must correctly utilize them together to produce an accurate deliverable.

The deliverable's use case demands a quality notion beyond computational accuracy.
To illustrate, consider the business function of this task.
For a fund, a modeling exercise like this answers a core capital allocation question: \textit{should one acquire Macrohard, and at what price?}
To that end, the fund cares about IRR, which determines whether the investment's return is high enough to be worthwhile. 
In practice, however, IRR depends on numerous modeling choices (e.g. whether to model growth of Macrohard's product lines separately) and forecast values (e.g. whether product A grows at 2\% or 5\%, and for how many years). 
After an analyst hands off the deliverable, these high-level assumptions are routinely contested by upper management, from the analyst's direct managers to VPs. 
Familiar with spreadsheets themselves, they often experiment on top of the deliverable with different assumptions to reach a decision on tight timelines.
As a result of this collaborative and high-stakes use of the deliverable, deriving a correct value or summary matters less than developing a readable and easily modifiable deliverable. 
We discuss a taxonomy that grounds these desiderata in concrete, evaluable criteria in Section~\ref{sec:eval_criteria}.

%% file: sections/04_eval_criteria.tex
To reflect the multifaceted aspects of quality necessary in spreadsheets for financial workflows, we propose a taxonomy organized around three core dimensions:  \textbf{Accuracy}, \textbf{Format}, and \textbf{Formula}. Each core dimension is composed of multiple sub-dimensions, as illustrated in Figure~\ref{fig:benchmark_results}.
At the high level: (a) \textbf{Accuracy} assesses whether the agent performs the desired task (e.g. did the agent perform scenario analysis) and whether it is done accurately, (b) \textbf{Formula} evaluates the robustness and usability of the functions that agents employ (e.g. would calculations break if assumptions change), and (c) \textbf{Format} measures the clarity and professionalism of the deliverable's presentation.

To illustrate, we take \texttt{Logic Readability \& Size} of the Formula dimension as an example.

\begin{figure}[ht]
    \centering
    \includegraphics[width=\textwidth]{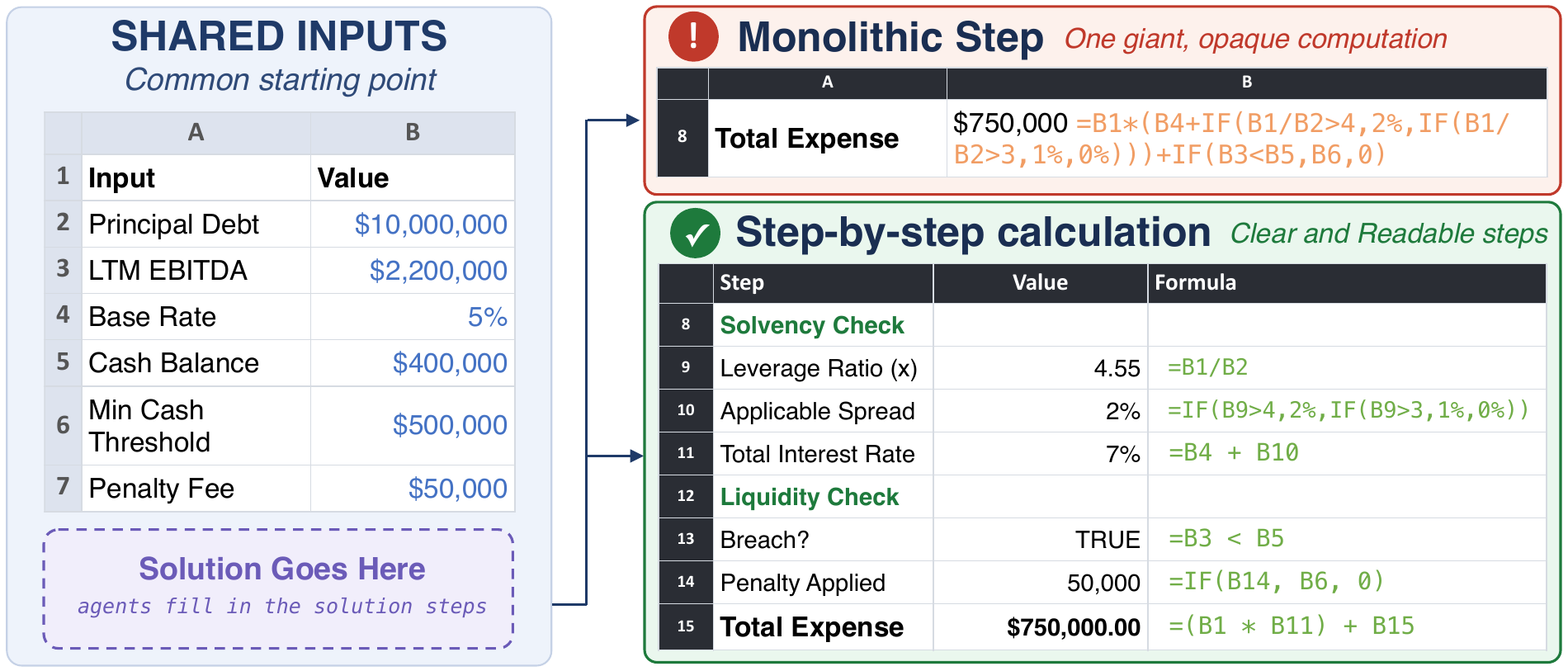}
    \caption{
        Comparison between a unreadable, monolithic function (\textit{top right}) and a clear, intent-revealing presentation of the calculation (\textit{bottom right}). 
        Although both arrives at the same result, only the latter is acceptable in professional setting. 
    }
    \label{fig:monolithic}
\end{figure}

\textbf{Formula --- Logic Readability \& Size.}
Consider the task of computing a company’s total expenses given its principal debt, interest rate, cash balance, and related quantities.
In principle, the entire computation can be performed in a single cell, as shown at the top right of Figure~\ref{fig:monolithic}.
However, the resulting formula is difficult to parse and to audit, especially for time-constrained managers who need to verify the calculation.
By contrast, the bottom right of Figure~\ref{fig:monolithic} decomposes the calculations in steps, presenting interpretable intermediate quantities that enable quick sanity checks and clearly expose how each contributes to the final total expense estimate.
Although both approaches yield the same numerical result, the decomposed version is clearly preferable in a professional setting.
We provide a description of all sub-dimensions, as well as additional examples, in Appendix~\ref{app:subdimensions}.

%% file: sections/05_setup.tex
\textbf{Task Collection.}
We curate a diverse set of end-to-end spreadsheet tasks in finance from three primary sources:
the Financial Modeling World Cup (FMWC), ModelOff, and the Wall Street Prep (WSP) training curriculum \citep{fmwc, modeloff, wsp}.
These sources are widely used in both professional and educational settings and cover a broad spectrum of core financial workflows from valuation to investment strategy.
Tasks from these platforms are designed to reflect real workplace expectations, requiring end-to-end construction of multi-sheet Excel workbooks.

\textbf{Task Annotations in Difficulties and Types.}
To better understand the tasks, we manually annotate the available tasks into different difficulty levels and types (see Figure~\ref{fig:diff_and_types}).
Typically, Level 2 (Easy) tasks are doable without deep financial expertise, whereas level 5 are difficult for even seasoned modelers.
The benchmark contains tasks spanning levels 1–5.
On task types (right of Figure~\ref{fig:diff_and_types}),
we find the tasks spanning a wide range of core financial workflows.
DCF, for example, is the most common method to value a company in finance.
We detail the annotation process in Appendix~\ref{app:Task Difficulty Classification}.

\textbf{Collecting Agent Attempts.}
Due to the growing number of proprietary agents \citep{claude_for_excel, chatgpt_for_excel, openai_chatgptagent} for spreadsheet tasks,
evaluating capable agents available to the public poses a unique challenge to benchmarks.
Proprietary products (\textbf{\guiagentname}) such as ChatGPT for Excel \cite{chatgpt_for_excel} are only available via GUI-based interactions, and do not expose accessible API endpoints.
To collect attempts from these agents, we develop a pipeline based on playwright \citep{playwright2025} that automatically interacts with the products to collect results.
Note that the pipeline itself can be of general interest for researchers who want to evaluate \guiagentname models.
We provide the code alongside the benchmark for future use.
Beyond the GUI-based products, frontier labs increasingly ship \textit{coding harnesses} (Codex \citep{openai_codex} and Claude Code \citep{claude_code}) that can operate the spreadsheet by writing and executing code to build the workbook.
We treat these as a third class, \textbf{\codeagentname} agents, and collect their attempts by running the corresponding agentic workflow in isolated docker containers.
Finally, to control for effects from differing agentic harnesses behind the proprietary solutions, we develop our own agentic framework to benchmark available LLMs on equal footing.
We refer to these agents as \textbf{\apiagentname} agents.
Because \guiagentname, \codeagentname, and \apiagentname agents can share an identical underlying model (e.g. Fable 5 powers Claude Cowork, Claude Code, and our own harness), results from these groups allow us to isolate the impact of agentic harnesses.
Additional details of the agents evaluated and the three pipelines are provided in Appendix~\ref{app:agents}.

\textbf{LLM-as-Judge Evaluation.}
As discussed in Section~\ref{sec:eval_criteria}, spreadsheet quality criteria are nuanced and often not amenable to exact-match verification (e.g. \texttt{Logic Readability \& Size} in Figure~\ref{fig:monolithic}).
We therefore adopt an LLM-as-judge evaluation pipeline
and analyze its reliability in Section~\ref{sec:judge_eval}.
The judge receives an evaluation rubric detailing the sub-dimensions criteria (Appendix~\ref{app:subdimensions}) along with structured representations of the reference solution and the agent's completed spreadsheet.
Then, the judge provides its assessment in a JSON format (see Figure~\ref{fig:judge-output}), containing a pass or fail verdict for each criterion, and when failing, a detailed account of where and why the attempt fell short.
In this paper, we primarily use the judge's verdict to grade attempts.
However, we note that the judge is capable of providing genuinely helpful findings that can be of independent interest (see examples in Section~\ref{sec:judge_eval}).
Throughout the paper, we use Gemini-2.5-pro as the backbone model \citep{comanici2025gemini25}

\begin{wrapfigure}{r}{0.40\columnwidth}
\vspace{\intextsep}  %
\begin{tcolorbox}[
    enhanced,
    colback=oxford-bg,
    colframe=oxford-header,
    boxrule=0.6pt,
    arc=2pt,
    width=\linewidth,
    top=1mm, bottom=1mm, left=1mm, right=1mm,
    title={\textsf{\textbf{LLM Judge Output}}},
    fonttitle=\small,
    coltitle=white,
    toptitle=1mm, bottomtitle=1mm,
]
\begin{lstlisting}[style=json,basicstyle=\footnotesize\ttfamily]
"Formulas": {
  "name": "Avoid hardcoding",
  "decision": "Fail",
  "mistakes": [{
    "description": "Hard-coded
      timeline headers.",
    "location": "Model!D64:AZ65",
    "explanation": "Year/Month
      headers are typed manually;
      should be computed from the
      Model Start Date input.",
    "expected": "Use YEAR(),
      MONTH(), DATE() to generate
      headers dynamically."}]}
\end{lstlisting}
\end{tcolorbox}
\caption{
An LLM judge output for the formula rubric.
Here, the judge correctly identifies that the entries are hardcoded, resulting in a brittle spreadsheet if model's start date changes.
Note that these mistakes would have otherwise been difficult to detect with exact-matching-based verification.
}
\vspace{-4em}
\label{fig:judge-output}
\end{wrapfigure}

\textbf{Criteria Weighting.}
To enable quantitative cross-model comparison, we assign weights to the evaluation dimensions and sub-dimensions reflecting their relative importance in finance practices (left of Figure~\ref{fig:benchmark_results}).
We acknowledge that any such weighting is inherently subjective, and it serves only to produce a concise scalar summary for model comparison.
Given the rich output from the LLM judge, one can easily adjust the weighting to derive composite scores suited to their settings.

%% file: sections/06_judge_eval.tex
We next analyze the behavior of the LLM judge.
To assess the robustness of the judge under natural variability in agents' attempts, we evaluate it on two complementary attempt types:

\textbf{(1) Synthetic perturbations:} we construct controlled variants of gold-standard solutions by injecting targeted errors
for each of the subdimensions. These perturbations enable precise testing of whether the judge correctly identifies minor errors for each subdimension.

\textbf{(2) LLM agent attempts:} we evaluate the judge on complete solutions produced by a diverse set of frontier LLM agents. These attempts represent real error patterns in current agents, which
may deviate substantially from the reference solution in structure and logic.
Note that regardless of these choices, so long as the deliverable satisfies the rubric, the attempt should be marked as correct.
Therefore, this evaluation provides an additional stress test of the judge’s ability to produce correct judgments and feedback.

Together, these two regimes probe complementary aspects of judge reliability.
We document the judge's performance on synthetic perturbations and agent attempts respectively in Tables~\ref{tab:perturbations} and \ref{tab:ai_attempts}.
As shown there, the judge decisions largely align with experts' judgment.
In particular, on grading agent attempts, the judge achieves a \judgeacc accuracy, \judgebacc balanced accuracy, and \judgefone $F_1$ score when measured against \annotcount expert annotations.

The judge also provides a rich text-based assessment of where the attempt falls short in the case of failures.
Here, we find surprisingly nuanced judgment that requires a deep and accurate understanding of spreadsheet workflows and the task context, and highlight a couple of notable cases below.

\textbf{Catching Off-by-One Error.} In one case where calculating the value of interest (Last Twelve Months EBITDA) requires aggregating revenues from four quarters prior for every quarter, the Claude Web agent \citep{claude_web} commits an off-by-one error in the range of windows aggregated. However, since the quarterly values are numerically close, the agent in fact accidentally arrives at the expected numerical value. As a result, a simple exact-match-based verification could not have caught this error. Nonetheless, the judge identified:
\begin{trajectory}
    "\textbf{decision:}" "Fail",

    "\textbf{summary}": "Model contains a logical error ... though the answers are coincidentally unaffected ...",

\end{trajectory}

\textbf{Catching Hardcoded Solutions.}
In a trading simulation task spanning 1{,}000+ rows, Claude Web~\citep{claude_web} populates the output columns with hardcoded values rather than formulas, bypassing Excel entirely by pasting in numerically correct but externally computed results.
A faithful implementation requires helper columns that decompose the simulation logic into traceable, auditable steps.
As in the previous case, the violation here is invisible to value-based verification: the failure is not a malformed formula but the absence of one. The judge identified:

\begin{trajectory}
    "\textbf{decision:}" "Fail",

    "\textbf{summary}": "Core logic is not implemented with formulas. Instead, the results of the simulation ... are hard-coded. This completely obscures the logic and makes the model unauditable.",
\end{trajectory}

Regarding expert effort, iterating on a clear description of criteria for judge's rubric, constructing the perturbation suite, and annotations of Tables~\ref{tab:perturbations} and \ref{tab:ai_attempts} require substantial expert involvement.
In total, 2 MBAs and 3 finance professionals spent \experthours hours.
We provide additional judge details and analysis in Appendix~\ref{app:judge}.

\begin{table*}[htbp]
\centering
\footnotesize
\setlength{\tabcolsep}{4pt}
\renewcommand{\arraystretch}{1.0}
\begin{tabular}{@{}ll@{\hspace{4pt}}cccc@{\hspace{8pt}}cccc@{\hspace{8pt}}cccc}
\toprule
 &  & \multicolumn{4}{c}{\textsc{Claude (Web)}} & \multicolumn{4}{c}{\textsc{Claude (Excel)}} & \multicolumn{4}{c}{\textsc{ChatGPT (Excel)}} \\
\cmidrule(lr){3-6} \cmidrule(lr){7-10} \cmidrule(lr){11-14}
\textsc{Category} & \textsc{Error Type} & 168 & 187 & 61 & 108 & 168 & 187 & 61 & 108 & 168 & 187 & 61 & 108 \\
\midrule
\multirow{4}{*}{{\textsc{Accuracy}}} & Final Calculations & \celltn & \cellyes & \celltn & \celltn & \cellno & \celltn & \cellno & \celltn & \cellyes & \cellfp & \cellyes & \celltn \\
 & Starting Values & \celltn & \celltn & \celltn & \celltn & \cellyes & \celltn & \cellyes & \celltn & \celltn & \celltn & \cellyes & \celltn \\
 & Task Completed & \celltn & \celltn & \celltn & \celltn & \cellyes & \celltn & \cellno & \celltn & \celltn & \celltn & \cellyes & \celltn \\
 & Number Sign & \celltn & \celltn & \celltn & \celltn & \celltn & \celltn & \cellyes & \celltn & \celltn & \celltn & \cellyes & \celltn \\
\midrule
\multirow{5}{*}{{\textsc{Formula}}} & Logic Readability & \cellyes & \celltn & \celltn & \cellyes & \celltn & \cellno & \cellyes & \celltn & \cellyes & \cellyes & \cellyes & \celltn \\
 & Edge Cases (\#DIV/0!) & \celltn & \celltn & \celltn & \cellyes & \cellyes & \celltn & \cellyes & \celltn & \celltn & \celltn & \cellyes & \celltn \\
 & Hardcoded values & \cellyes & \cellyes & \celltn & \cellyes & \cellyes & \cellyes & \cellyes & \celltn & \cellyes & \cellyes & \cellyes & \celltn \\
 & Range Issues & \cellyes & \celltn & \cellyes & \cellyes & \celltn & \celltn & \cellyes & \cellyes & \celltn & \celltn & \cellyes & \cellyes \\
 & Absolute References & \celltn & \celltn & \celltn & \celltn & \celltn & \celltn & \cellyes & \celltn & \cellyes & \celltn & \celltn & \celltn \\
\midrule
\multirow{8}{*}{{\textsc{Format}}} & Sheet Structure & \celltn & \celltn & \celltn & \celltn & \cellno & \cellyes & \cellno & \celltn & \celltn & \celltn & \celltn & \celltn \\
 & Readability & \celltn & \cellyes & \celltn & \cellno & \cellyes & \cellno & \cellno & \cellno & \cellyes & \cellyes & \celltn & \cellno \\
 & Color Scheme & \cellno & \celltn & \celltn & \cellyes & \cellyes & \cellyes & \cellyes & \cellyes & \cellyes & \cellyes & \cellyes & \celltn \\
 & Number Notation & \celltn & \cellfp & \celltn & \cellno & \cellyes & \cellyes & \cellyes & \cellyes & \celltn & \celltn & \cellyes & \cellno \\
 & Alignment & \celltn & \celltn & \celltn & \celltn & \celltn & \celltn & \cellyes & \celltn & \celltn & \celltn & \cellno & \cellno \\
 & Font Size\ /\ Style & \cellyes & \cellfp & \celltn & \cellyes & \celltn & \cellyes & \celltn & \cellyes & \cellyes & \celltn & \celltn & \cellyes \\
 & Borders & \cellyes & \cellyes & \celltn & \cellyes & \celltn & \cellno & \celltn & \cellyes & \cellyes & \cellno & \celltn & \cellyes \\
 & Output Presentation & \celltn & \celltn & \celltn & \celltn & \celltn & \celltn & \celltn & \celltn & \celltn & \celltn & \celltn & \celltn \\
\bottomrule
\end{tabular}
\caption{
    The judge's performance on grading agent attempts.
    \colorbox{lightblue}{\makebox[1em]\checkmark} = error caught by the judge,
    \colorbox{lightpurple}{\makebox[1em]\checkmark} = correct solution recognized by the judge,
    \colorbox{lightred}{\makebox[1em]{$\times$}} = error not caught,
    \colorbox{lightamber}{\makebox[1em]{$\times$}} = error flagged but not present.
}
\label{tab:ai_attempts}
\end{table*}

%% file: sections/07_results.tex
\textbf{Agent Performance from Expert Inspection.}
Inspecting the agents' attempts reveals a surprising pattern: despite being released for spreadsheet workflow, Claude (Excel) \citep{claude_for_excel} and ChatGPT (Excel) \citep{chatgpt_for_excel} trail Claude Web \citep{claude_web} in spreadsheet quality.
Claude Web produces consistently cleaner formatting, whereas the Excel agents frequently introduce errors such as inconsistent font colors and missing unit labels.
Notably, the Excel agents and ChatGPT Pro \citep{chatgpt_web} tend to hardcode values rather than adding the underlying calculations as Excel formulas, yielding workbooks that are professionally inadequate and difficult to edit.
The hardcoded values themselves are typically correct, suggesting that these agents can carry out the calculations in some other environment but fail to express them natively on the spreadsheet.

Holding the underlying frontier model fixed (e.g. Fable 5 and GPT-5.6 Sol), \guiagentname and \codeagentname agents on proprietary harnesses substantially outperform \apiagentname agents on our in-house agentic harness, indicating that harness quality drives a large portion of the gap.
We provide additional per-agent analysis in Appendix~\ref{app:agent_analysis}.

\begin{figure}[h]
    \centering
    \includegraphics[width=\textwidth]{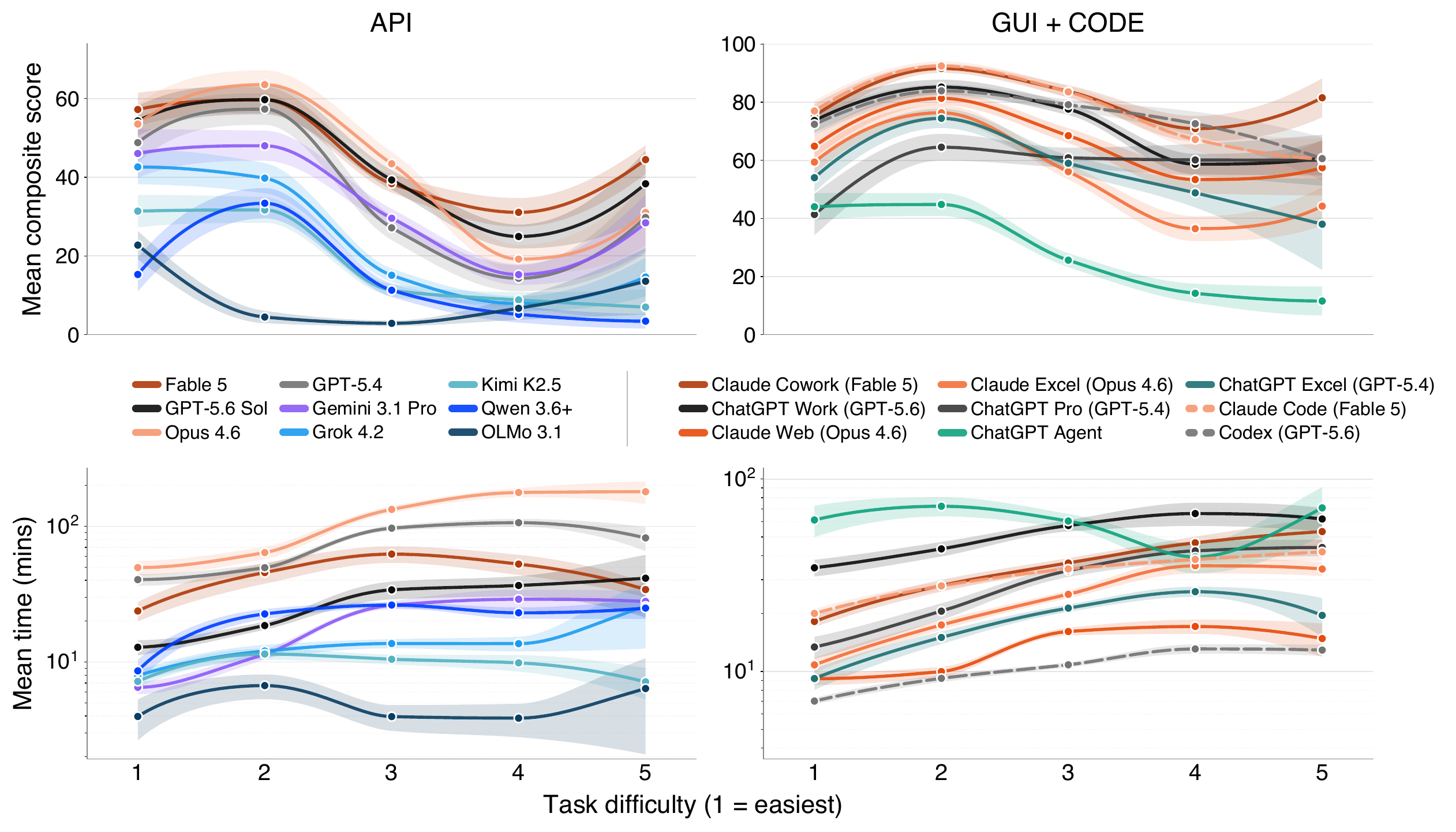}
    \caption{
        Mean attempt completion time (\textit{bottom}) and mean composite score (\textit{top}) plotted against task difficulty. (\textit{Left}) \apiagentname agents. (\textit{Right}) \guiagentname and \codeagentname agents.
        Agents seem to recognize task difficulty, and generally spend more time as difficulty increases.
        However, their performance nonetheless degrade significantly, suggesting that harder spreadsheet tasks in \benchmarkname poses difficulty that test-time scaling struggle to overcome.
        Interestingly, agents (e.g. ChatGPT Agent, Kimi K2) that do not increase attempt time as difficulty increases also perform worse on \benchmarkname.
        Shaded area indicates standard error.
    }
    \label{fig:agent_time}
\end{figure}

\textbf{Aggregate Results.}
We report benchmark results for the evaluated agents in Figure~\ref{fig:benchmark_results}.
Many quantitative results therein align with our qualitative observations:
(1) except ChatGPT Agent, state-of-the-art agents deployed on proprietary harnesses lead all others by a clear margin, with the Fable 5 agents performing the best, followed closely by the GPT-5.6 Sol agents;
(2) the Claude family of agents appears the strongest in spreadsheet manipulation; and
(3) agents based on the in-house harness rank well below this leading group.

Beyond the top-performing \guiagentname and \codeagentname agents, our evaluation also exposes finer-grained capability differences among \apiagentname agents.
In particular, leading SOTA models such as Fable, Opus, and GPT Sol still outperform alternatives, followed by Gemini 3.1 Pro, Grok 4.2, and the remaining models, with a clear margin between these successive tiers.
The clear margins show that our agentic framework is able to differentiate between models of varying capabilities, and can serve as a useful tool for model evaluation and comparison in the spreadsheet domain.
Interestingly, the performance of leading models do not differ significantly under our agentic framework.
Considering that these same models are more clearly separated (e.g. Claude Web Opus 4.6 trails behind the top agents) on proprietary harnesses, points to the fact that harness rather than raw model capability is a main driver of end-to-end spreadsheet performance.

\textbf{Sensitivity to Difficulty.} To better understand the agent's performance on varying degrees of task difficulty, we analyze the composite score as a function of the difficulty annotations described in Section~\ref{sec:benchmark}.
The top portion of Figure~\ref{fig:agent_time} shows that for both \apiagentname agents (left) and \guiagentname/\codeagentname agents (right),
there exists a trend of performance degradation as difficulty increases, with the top agents scoring only 72.6 out of 100 at level 4 tasks (Codex), dropping from 92.4 at level 2 (Claude Code).
Recall that tasks at level 2 difficulty do not require deep financial expertise, and hence it is not surprising that many commercial products are already capable of completing them to a satisfactory degree.

Next, we examine whether the time it takes for agents to complete an attempt changes as a function of difficulty.
We also observe here that most agents appear to recognize the increase difficulty by allocating more time and compute.
Note that there are notable exceptions (e.g. Qwen, Kimi, and ChatGPT Agent) that do not seem to respond to increased difficulty, and spend roughly the same amount of time on tasks across difficulty.
Interestingly, these are also the agents that underperforms others in the same class as shown in Figure~\ref{fig:benchmark_results}, suggesting that the underlying models may be incapable of comprehending spreadsheet difficulty.
Finally, we note while Claude Web (Opus 4.6) is among the fastest proprietary agents, Opus costs the most time among all \apiagentname agents; this points to harness design as a key factor driving efficiency of the agents, besides clear benefit to performance.

Overall, the benchmark reveals substantial and systematic quality gaps both across and within harness families.
Even the strongest agent, Claude Cowork, is far from reliably achieving all professional standards expected from experienced models, and all agents experience a notable performance drop as task difficulty increases: on Medium-Hard (level 4) tasks, the best agent is Codex at only 72.6 out of 100.
The results underscores the significant headroom that remains before LLM agents can reliably produce financial spreadsheets at a professional standard.

%% file: sections/02_related_work.tex
\textbf{Beyond Exact Verifiers.}
A broad range of benchmarks have been proposed to evaluate LLM-based agents across diverse domains, including software engineering \citep{jimenez2024swebench, yang2024sweagent}, 
web and GUI interaction \citep{zhou2023webarena, osworld_verified, chezelles2025browsergym}, 
and tool and API use \citep{guo2024stabletoolbench, patil2025bfcl}. 
Despite their diversity, these benchmarks all evaluate based on exact matching of certain solution or environment end states (e.g., passing unit tests for code). 
While extremely successful as goalposts, 
the focus on exactly verifiable evaluation necessarily flattens the nuanced sense of task quality prevalent in real-world tasks. 
As an example, the financial spreadsheet tasks require a rich quality standard that cannot be reduced to matching a pre-defined string or artifact state.
Our benchmark, therefore, departs from the exact-match-based evaluation paradigm of prior agent benchmarks. 
In this sense, our work is most similar to GDPval \citep{patwardhan2025gdpval_evaluatingaimodel}, which also explores agent's performance on real-world tasks.
To tackle the inherent difficulty in assessing task quality, GDPval employs experienced professionals to compare solutions for evaluation. 
In contrast, we provide a scalable and transparent evaluation using a validated LLM-based judge.

\textbf{Benchmarking LLM on Spreadsheet in Finance.}
Spreadsheets are the lingua franca of financial workflows. 
Within this scope, 
SpreadsheetBench \citep{spreadsheetbench} has emerged as the de facto standard benchmark, cited in the releases of both Opus 4.5 and ChatGPT Agent \citep{anthropic2025claudeopus45, openai_chatgptagent}.
However, the tasks therein focus on simple atomic tasks centered on introducing correct functions or making the expected edits on predetermined cells. 
Similarly to common ML benchmarks, prior work on evaluating LLMs in the financial domain has largely focused on isolated question answering and blank filling rather than end-to-end financial task completion. 
Numerous benchmarks such as PIXIU \citep{xie2023pixiu}, FinBen \citep{FinBen}, BizBench \citep{krumdick-bizbench}, BizFinBench \citep{lu2025bizfinbench} and XFinBench \citep{XFinBench} primarily evaluate models on financial QA, numerical reasoning, or short-form problem solving.
In contrast, similar to GDPval \citep{patwardhan2025gdpval_evaluatingaimodel} our work targets full financial modeling workflows in spreadsheets and introduces a holistic taxonomy of solution quality that encompasses not only numerical accuracy but also structural organization and formula transparency, reflecting professional standards of usability. 
Consequently, this work is among the first to systematically propose and operationalize such multi-dimensional criteria for evaluating agents on end-to-end financial spreadsheet tasks.
To this end, we explore and validate an LLM-based judge that performs this inherently ambiguous, difficult to verify, but nonetheless valuable artifact-level evaluation.

%% file: sections/08_conclusion.tex
\benchmarkname evaluates LLM agents on end-to-end spreadsheet tasks,
moving beyond previously studied atomic tasks such as question answering or summarization.
To formalize task quality in this richer setting, we propose a taxonomy organized around three core dimensions
(\textbf{Accuracy}, \textbf{Formula}, and \textbf{Format}), each decomposed into
expert-informed subdimensions that mirror professional standards in finance.
We operationalize the taxonomy through an LLM-based judge, which we then validate against expert
annotations on both controlled synthetic perturbations and a diverse set of agent attempts spanning \annotcount annotations.

Our results show that state-of-the-art models can already complete easier end-to-end financial tasks at a reasonable quality, but substantial headroom for improvement remains, especially for harder tasks that require longer chains of calculations.
Among the evaluated agents, we find that those deployed on the proprietary graphical and coding harnesses shipped by frontier labs lead all others by a clear margin.
The results of agent performance under in-house and proprietary harnesses point to agentic harnesses being the driver factor in performance and efficiency in spreadsheet tasks.

In terms of limitations, \benchmarkname is focused on end-to-end spreadsheet tasks in finance.
As a result, while many criteria studied are general good practices in spreadsheets and the LLM evaluation pipeline is fully transferable to other rubrics, some criteria can be finance-specific (e.g. \texttt{Number Notation}).
The benchmark targets workflows such as financial modeling, forecasting, valuation,
scenario analysis, and related spreadsheet-based analyses.
Moreover, as we lack visibility in the proprietary agentic harnesses that frontier labs employ, we cannot conclusively determine the primary source of the differences in agents' performance observed.
Finally, while we observe notable performance drop in all agents as difficulty increases, the benchmark does not have a substantial number of the hardest tasks in levels 5 and 6 according to our categorization.
We leave the development of harder spreadsheet tasks as an interesting and important future direction.

\section*{Acknowledgements}
The authors thank the Laude Institute for their generous support of this work.